 \RequirePackage{fix-cm}
 \documentclass[smallextended]{svjour3}       
 \smartqed  
 \usepackage{amssymb}
 \usepackage{float}
 \usepackage{caption}
 \usepackage{subcaption}
 \usepackage{graphicx}
 \usepackage{verbatim}
 \usepackage{epstopdf}
 \usepackage{amsmath}
 \usepackage{mathtools}
 \usepackage{epstopdf}
\pdfoutput=1
 
 \usepackage{gensymb}
 \usepackage[]{algorithm2e}
 \usepackage[justification=centering]{caption}

 \graphicspath{{Images/}}
 \usepackage{graphicx}
 \usepackage{natbib}
 \setcitestyle{authoryear,open={(},close={)}}

\begin{document}
 	
 	\title{On Efficient Global Optimization via Universal Kriging Surrogate Models
 	}
 	
 	
 	\author{Pramudita Satria Palar        \and
 		Koji Shimoyama 
 	}
 	
 	
 	\institute{F. Author \at
 		Institute of Fluid Science, Tohoku University, Sendai, Miyagi Prefecture 980-8577, Japan \\
 		\email{palar@edge.ifs.tohoku.ac.jp}           
 		\and
 		S. Author \at
 		Institute of Fluid Science, Tohoku University, Sendai, Miyagi Prefecture 980-8577, Japan \\
 		\email{shimoyama@edge.ifs.tohoku.ac.jp} 
 	}
 	
 	\date{Received: date / Accepted: date}

 	\maketitle
 	
 	\begin{abstract}
 		In this paper, we investigate the capability of the universal Kriging (UK) model for single-objective global optimization applied within an efficient global optimization (EGO) framework. We implemented this combined UK-EGO framework and studied four variants of the UK methods, that is, a UK with a first-order polynomial, a UK with a second-order polynomial, a blind Kriging (BK) implementation from the ooDACE toolbox, and a polynomial-chaos Kriging (PCK) implementation. The UK-EGO framework with automatic trend function selection derived from the BK and PCK models works by building a UK surrogate model and then performing optimizations via expected improvement criteria on the Kriging model with the lowest leave-one-out cross-validation error. Next, we studied and compared the UK-EGO variants and standard EGO using five synthetic test functions and one aerodynamic problem. Our results show that the proper choice for the trend function through automatic feature selection can improve the optimization performance of UK-EGO relative to EGO. From our results, we found that PCK-EGO was the best variant, as it had more robust performance as compared to the rest of the UK-EGO schemes; however, total-order expansion should be used to generate the candidate trend function set for high-dimensional problems. Note that, for some test functions, the UK with predetermined polynomial trend functions performed better than that of BK and PCK, indicating that the use of automatic trend function selection does not always lead to the best quality solutions. We also found that although some variants of UK are not as globally accurate as the ordinary Kriging (OK), they can still identify better-optimized solutions due to the addition of the trend function, which helps the optimizer locate the global optimum.

 	\end{abstract}
 	
 	\section{Introduction}
 	Recent rapid advances in computing power and technology have fostered an increased use of computational optimization methods in solving engineering design optimization problems. The typical goal of engineering design optimization is to maximize the performance of an engineering system thereby achieving the optimum output. Design optimization might also reveal important information that is useful in the overall design process. In spite of such advancements, real-world computationally expensive optimization problems remain rather difficult, if not impossible, to be solved via traditional optimization methods. This difficulty might be caused by the nonavailability of gradient information, high computational costs, or failed simulations that return no results. To overcome these challenges, one must resort to modern and advanced optimization methods that are specifically designed to solve  real-world problems. Here, the surrogate model is an invaluable tool that assists in optimization when one function evaluation is computationally expensive. In engineering design optimization, a surrogate model is frequently used for global optimization, in which the goal is to find the global optimum of a given optimization problem. Among surrogate models, the Kriging model~\citep{krige1951statistical, sacks1989design, simpson2001kriging} is arguably one of the most popular surrogate models for optimization due to its interpolation capability and flexibility. The versatility of the Kriging method has fostered a wide use of the method in solving numerous many real-world problems not just limited to optimization. More specifically, Kriging has been used for aerodynamic optimization~\citep{jeong2005efficient}, structural optimization~\citep{sakata2003structural, forsberg2005polynomial}, and uncertainty quantification~\citep{dwight2009efficient, shimoyama2013dynamic}, to name a few. In this paper, we specifically focus on the application of Kriging for optimization; however, we note that Kriging is also applicable for uncertainty quantification problems that involve optimization, such as the quantification of upper and lower bounds of output uncertainty as presented by~\cite{swiler2009epistemic}.
 	
 	One advantage of Kriging is that it directly provides the estimation error; very few surrogate models offer estimation error. Estimation errors are useful in achieving such goals as refining the surrogate model's global accuracy or efficiently guiding the optimization process that balances exploration and exploitation. The efficient global optimization (EGO) framework solves optimization problems by using a metric called expected improvement (EI) that relies on both prediction and error measures provided by the Kriging model~\citep{jones1998efficient}. Recent research involving EGO-based methods includes bootstrapped Kriging that implements EGO using an improved estimator of the Kriging predictor variance through bootstrapping~\citep{kleijnen2012expected} and an EGO implementation with a fully Bayesian approach~\citep{benassi2011robust}. Moreover, an EI-based model equipped with constraint handling was proposed to efficiently handle constrained problems~\citep{parr2012infill}. Further, aside from single-objective EGO, multiobjective EGO implementations are also available~\citep{jeong2005efficientb, knowles2006parego, keane2006statistical}. Here, the underlying principle of  multiobjective EGO implementations is similar to that of single-objective EGO, with some modifications required such that it can be used to handle multiobjective optimization problems. Apart from deterministic optimization, the EGO framework has also been refactored such that it can solve robust optimization problems~\citep{ur2014efficient, ur2015efficient}. EGO is also applicable to parallel optimization problems with such modifications as the use of multiple surrogates~\citep{viana2013efficient}.
 	
 	From the viewpoint of the Kriging model itself, the original EGO framework and its variants frequently use OK as their backbone. Universal Kriging (UK)~\citep{matheron1969cahiers} has yet to become a widely accepted tool for optimization because of its difficulty predicting the underlying trend in the surface to be approximated~\citep{kersaudy2015new}. To address this problem, methods that automatically select a suitable trend function have been proposed. Among the first of these is the blind Kriging (BK) method, which performs Bayesian forward selection to identify a set of trend functions that can increase the approximation capabilities of Kriging~\citep{joseph2008blind}. BK tries to avoid an improper polynomial trend function that might result in disastrous approximations. Further, BK has been successfully implemented and shown to perform better than OK in such applications as turbomachinery design~\citep{bellary2016comparative}. As studied by~\cite{joseph2008blind} and \cite{couckuyt2012blind}, BK also showed better performance versus that of OK in several computer experiments, such as piston slap and sealing experiments. Finally, BK has been implemented in MATLAB as the ooDACE toolbox and can be downloaded online~\citep{couckuyt2014oodace}; however, note that the performance of BK shows only minimal improvements or none at all versus that of OK if enough data is available to cover the problem domain~\citep{couckuyt2012blind}. In addition to BK, trend function selection methods using a genetic algorithm (GA) have been proposed~\citep{zhao2011metamodeling, liang2014using}; such approaches have been called dynamic Kriging. These dynamic Kriging methods originally use Kriging variance as their objective functions~\citep{zhao2011metamodeling}, but such approaches have been criticized because the magnitude of Kriging variance decreases linearly with the number of trend functions included (i.e., under the assumption of constant hyperparameters)~\citep{liang2013comment}. Note that dynamic Kriging with error estimations predicted using cross-validation was proposed to overcome this problem~\citep{liang2014using}. Aside from the standard monic polynomial form, orthogonal polynomials can also be used as trend function for UK. As an example, a UK method with polynomial chaos expansion (PCE)~\citep{wiener1938homogeneous, xiu2002wiener} as the trend function was developed for uncertainty quantification and sensitivity analysis~\citep{schobi2015polynomial, kersaudy2015new}. The polynomial-chaos-Kriging (PCK) method is an algorithm that combines the capabilities of PCE and Kriging within the framework of UK. Instead of using a heuristic optimization procedure such as GA, PCK uses the least-angle-regression (LARS) algorithm to choose the most influential orthogonal polynomial function set. There are two distinct ways to implement PCK that is, either sequential or optimal-PCK, where the former builds the PCE and UK sequentially and the latter builds a new UK model at each iteration of the LARS algorithm. PCK has been applied to various engineering problems, including structural reliability problems and rare event estimations~\citep{schobi2014combining, schobi2016rare}. The success of BK and the PCK algorithm in creating a metamodel has motivated us to investigate their capabilities when coupled with the EGO framework to solve expensive optimization problems. Although BK has been used before for optimization, it is not yet implemented within the iterative global optimization strategy of EGO. The goal of our investigation is to further encourage the application of UK in solving expensive real-world optimization problems.  
 	
 	In this paper, we propose an EGO method based on the UK method with a polynomial trend; we call our combined approach the UK-EGO algorithm. Our method is a further extension and formal implementation of the UK method with a single-objective EGO framework. Similar to original EGO, we utilize the EI metric to guide the optimization process. Here, EI strikes a balance between exploration and exploitation of the design space; however, rather than directly using a constant trend function as in EGO, in our UK-EGO approach, we first exploit the possible trend in the response surface before applying EI-based optimization. The main contribution of our paper lies in the use of UK for EGO. More specifically, we study and compare the performance of BK from the ooDACE toolbox~\citep{couckuyt2014oodace}, PCK, and UK with fixed first- and second-order polynomial sets, all applied to the UK-EGO framework on synthetic and real-world test problems. 
 	
 	In addition to this introductory section, we structure the remainder of our paper as follows: in Section~\ref{sec:2}, we explain the fundamentals of the UK surrogate model; in Section~\ref{sec:3}, we provide details and the framework of our proposed method; in Section~\ref{sec:4}, we present our computational results for algebraic test problems and a real-world aerodynamic problem; finally, we provide our conclusions and suggestions for future work in Section~\ref{sec:5}.

 	\section{Universal Kriging surrogate model}
 	\label{sec:2}
 	In this section, we describe the fundamentals of the UK surrogate model; our discussion is focused more on explaining the UK. We also explain UK with automatic trend function selection (i.e., BK and PCK). For the following explanation of Kriging surrogate model, we primarily refer to the works of~\cite{sacks1989design} and~\cite{jones2001taxonomy}.
 	
 	\subsection{Basics}
 	 	\label{sec:2.1}
 	The goal of building a UK surrogate model within an optimization context is to approximate the relationship between the input $\boldsymbol{x}=\{x_{1},x_{2},\ldots,x_{m}\}^{T}$, where $m$ is the dimensionality of the decision variables, and output of interest $y=f(\boldsymbol{x})$ as a realization of the stationary Gaussian process $Y(\boldsymbol{x})$. This can be obtained by approximating the true function with the following UK model:
 	\begin{equation}
 	\label{eq:UK}
 	Y(\boldsymbol{x}) = \sum_{i=0}^{P-1}\alpha_{i}\Psi_{i}(\boldsymbol{x}) + Z(\boldsymbol{x}),
 	\end{equation}
 	where $\boldsymbol{\Psi}(\boldsymbol{x})=\{\Psi_{0}(\boldsymbol{x}),\ldots,\Psi_{P-1}(\boldsymbol{x})\}^{T}$ is a collection of regression, or trend, functions (usually polynomials), $\boldsymbol{\alpha}=\{\alpha_{0}(\boldsymbol{x}),\ldots,\alpha_{P-1}(\boldsymbol{x}) \}^{T}$ is the vector of corresponding regression coefficients, and $Z$ is a stochastic process that models the deviation from the global model. If the regression function is a constant, then the given UK simply becomes an OK. 
 	
 	The Kriging model assumes that a slight difference between the locations of two points corresponds to a small difference between their objective functions. This assumption is modeled by a statistical correlation between the two sets of variables. Although several correlation models are available, in this paper, we model the correlation between $Z(\boldsymbol{x}^{(i)})$ and $Z(\boldsymbol{x}^{(j)})$ using the Gaussian autocorrelation function as follows:
 	\begin{equation}
 	\boldsymbol{R}_{ij} = \text{corr}[Z(\boldsymbol{x}^{(i)}), Z(\boldsymbol{x}^{(j)})] = \text{exp}\bigg(-\sum_{k=1}^{m}\theta_{k}|x_{k}^{(i)}-x_{k}^{(j)}|^{2}\bigg),
 	\end{equation}
 	where $\boldsymbol{\theta}=\{\theta_{1},\ldots,\theta_{m} \}$ is the vector of hyperparameters to be estimated. Note that exponent 2 in the right-hand side term can also be set as one of the tunable hyperparameters; however, for simplicity, we assume that the value of this exponent is constant. Besides the Gaussian autocorrelation function, other types of correlation function can also be employed for constructing a Kriging model. For example,~\cite{stein2012interpolation} recommends the use of a Mat{\'e}rn class function instead of a Gaussian autocorrelation function, since Gaussian autocorrelation function makes a strong smoothness assumption, which might be unrealistic for real-world processes. However, in this paper, we use a Gaussian autocorrelation function since it is probably the most widely applied autocorrelation function for the construction of Kriging models within the context of engineering design optimization. 
 	
 	The approximation starts by collecting $n$ observations in the design space $\boldsymbol{\mathcal{X}}=\{\boldsymbol{x}^{(1)},\ldots,\boldsymbol{x}^{(n)} \}^{T}$ and their corresponding responses $\boldsymbol{y}=\{y^{(1)},\ldots,y^{(n)}\}^{T}=\{f(\boldsymbol{x}^{(1)}),\ldots,f(\boldsymbol{x}^{(n)}) \}^{T}$ to form the experimental design (ED). The UK predictor  is defined as
 	\begin{equation}
 	\hat{f}(\boldsymbol{x}) = \boldsymbol{\Psi}(\boldsymbol{x})^{T} \boldsymbol{\alpha} + \boldsymbol{r}(\boldsymbol{x})^{T}\boldsymbol{R}^{-1}(\boldsymbol{y}-\boldsymbol{F}\boldsymbol{\alpha}),
 	\end{equation}
	with the mean-squared error of the Kriging prediction $\hat{s}^{2}(\boldsymbol{x})$ calculated as	
 	\begin{multline}
 	\hat{s}^{2}(\boldsymbol{x}) = \sigma^{2}\big(1-(\boldsymbol{r}^{T}(\boldsymbol{x})\boldsymbol{R}^{-1}\boldsymbol{r}(\boldsymbol{x}))
 	+\big(\boldsymbol{F}^{T}\boldsymbol{R}^{-1}\boldsymbol{r}(\boldsymbol{x})-\boldsymbol{\Psi}(\boldsymbol{x})\big)^{T} \big(\boldsymbol{F}^{T}\boldsymbol{R}^{-1}\boldsymbol{F}\big)^{-1}\\
 	\big(\boldsymbol{F}^{T}\boldsymbol{R}^{-1}\boldsymbol{r}(\boldsymbol{x})-\boldsymbol{\Psi}(\boldsymbol{x}) \big)
 	\big).
 	\end{multline}
 	Here, $\boldsymbol{R}$ is an $n\times n$ matrix with $(i,j)$ entry as $\text{corr}[Z(\boldsymbol{x}^{(i)}), Z(\boldsymbol{x}^{(j)})]$ , $\boldsymbol{r}(\boldsymbol{x})$ is the correlation vector between $\boldsymbol{x}$ and $\boldsymbol{\mathcal{X}}$ with $(i,1)$ entry as $\text{corr}[Z(\boldsymbol{x}^{(i)}), Z(\boldsymbol{x})]$, and $\boldsymbol{F}=\{\boldsymbol{\Psi}(\boldsymbol{x}^{(1)}),\ldots,\boldsymbol{\Psi}(\boldsymbol{x}^{(n)}) \}^{T}$ is the $n\times P$ size matrix of regression functions. Coefficients $\boldsymbol{\alpha}$ are obtained by the generalized least-squares (GLS) procedure defined as
 	\begin{equation}
 	\label{eq:ls}
 	\boldsymbol{\alpha}=(\boldsymbol{F}^{T} \boldsymbol{R}^{-1}\boldsymbol{F})^{-1}\boldsymbol{F}^{T} \boldsymbol{R}^{-1}\boldsymbol{y}.
 	\end{equation}	
 	To calibrate the Kriging model, hyperparameters $\boldsymbol{\theta}$ must be estimated (denoted as $\boldsymbol{\hat{\theta}}$), which can be achieved by maximizing the likelihood function	
 	\begin{equation}
 	\label{eq:likelihood}
 	L(\boldsymbol{\alpha},\sigma^{2},\boldsymbol{\theta}) = \frac{1}{\sqrt{(2\pi\sigma^{2})^{n/2}|\boldsymbol{R}(\boldsymbol{\theta})|}}\text{exp}\bigg(-\frac{1}{2} \frac{(\boldsymbol{y}-\boldsymbol{F}\boldsymbol{\alpha})^{T}\boldsymbol{R}(\boldsymbol{\theta})^{-1}(\boldsymbol{y}-\boldsymbol{F}\boldsymbol{\alpha})}{\sigma^{2}} \bigg),
 	\end{equation}	
 	where the optimal estimate of $\boldsymbol{\alpha}$ is obtained through the least squares procedure, as in Equation~\ref{eq:ls}. The maximum likelihood estimates of the Kriging variance $\hat{\sigma}^{2}$ is	
 	\begin{equation}
 	\hat{\sigma}^{2}(\boldsymbol{\theta}) = \frac{1}{n}(\boldsymbol{y}-\boldsymbol{F}\boldsymbol{\alpha})^{T}\boldsymbol{R}(\boldsymbol{\theta})^{-1}(\boldsymbol{y}-\boldsymbol{F}\boldsymbol{\alpha}).
 	\end{equation}	
 	The likelihood function can be further simplified by substituting Equation~\ref{eq:ls} into Equation~\ref{eq:likelihood} and taking the natural logarithm of both sides. The simplified likelihood function is then defined as
 	\begin{equation}
 	\text{ln}(	L(\boldsymbol{\alpha},\hat{\sigma}^{2},\boldsymbol{\theta})) \approx -n \mbox{ ln} (\hat{\sigma}^{2}(\boldsymbol{\theta})) - \mbox{ln }(|\boldsymbol{R}(\boldsymbol{\theta})|).
 	\end{equation}
 	
 	Optimizing the likelihood function is difficult; therefore, a global optimizer such as a GA followed by the local search such as a Broyden-Fletcher-Goldfarb-Shanno (BFGS) is typically used. In this paper, we set the bounds of $\boldsymbol{\theta}$ for optimizing the likelihood function to $[10^{-3},10^{3}]$ with a logarithmic scale. 
 	
 	In practice, we always normalize the decision space to [-1,1] so that the Legendre polynomials can be used as trend functions. On the other hand, we normalize the function value according to 
 	\begin{equation}
 	\label{eq:krignor}
 	\tilde{y} = \frac{y-\mu({\boldsymbol{y}})}{\sigma({\boldsymbol{y}})},
 	\end{equation}
 	where $\mu({\boldsymbol{y}})$ and $\sigma({\boldsymbol{y}})$ are the mean and standard deviation of the function values of the current ED, respectively.
 	
 	\subsection{Automatic trend function selection for universal Kriging}
 	The most frequently used type of functions for UK approximation are polynomials. More specifically, multivariate polynomials from one-dimensional monic polynomials are widely used. In general, the trend function should be properly selected to create an accurate UK approximation, which is crucial, since the wrong/incorrect choice could result in  disastrously inaccurate approximations. Another polynomial form that can be used is the orthogonal polynomial, as used in the PCK approximation. Below, we describe the BK~\citep{joseph2008blind} and PCK methods~\citep{schobi2015polynomial,kersaudy2015new}, which both use an automatic trend function selection procedure to determine the most important polynomial terms given the current ED. We start by explaining the types of polynomials that can be employed for UK approximation. 
 	
 	\subsubsection{Choice of polynomials}
 	In this paper, we use the standard monic and Legendre polynomials as trend functions for BK and PCK, respectively. The monic polynomials used in BK (denoted as $\phi(x)$) are nonorthogonal, whereas the PCK approximation employs the orthogonal Legendre polynomials. The Legendre polynomials are a sequence of polynomials that are orthogonal with respect to the $L^{2}$ inner product on interval $[-1,1]$, mathematically defined as follows	
 	\begin{equation}
 	\int_{-1}^{1} \psi_{i}(x)\psi_{j}(x)dx = \frac{2}{2i+1}\delta_{ij},
 	\end{equation} 	
 	where $\psi(x)$ is the one-dimensional Legendre polynomial, and $\delta_{ij}$ is 1 if $i=j$ and 0 if $i\neq j$.
 	
 	Legendre polynomials can also be obtained via the Gram-Schmidt process on monic polynomials with respect to the $L^{2}$ inner product. Table~\ref{table:basisseq} depicts the first few Legendre polynomials in comparison with monic polynomials in the $[-1,1]$ domain.
 		
 	\begin{table}[h]
 	 		\caption{Sequence for monic and Legendre polynomials up to the fifth order.}
 	 		\label{table:basisseq}
 	 		\centering
 	 		\begin{tabular}{lll} 
 	 			\hline\noalign{\smallskip}
 	 			$p$  & Monic ($\phi(\boldsymbol{x})$) & Legendre ($\psi(\boldsymbol{x})$)\\ \hline
 	 			\noalign{\smallskip}\hline\noalign{\smallskip}
 	 			0 & 1 & 1 \\
 	 			1 &  $x$ & $x$ \\
 	 			2 & $x^{2}$ & $\frac{1}{2}(3x^{2}-1)$ \\
 	 			3 & $x^{3}$ & $\frac{1}{2}(5x^{3}-3x)$ \\
 	 			4 & $x^{4}$ & $\frac{1}{8}(35x^{4}-30x^{2}+3)$ \\ 
 	 			5 & $x^{5}$ & $\frac{1}{8}(63x^{5}-70x^{3}+15x)$ \\ \noalign{\smallskip}\hline
 	 		\end{tabular}
 	 		
 	 	\end{table}
 	
	Consider the index set $\boldsymbol{\zeta} = \{\zeta_{1},\ldots,\zeta_{m}\}$, where $\zeta_{i}=0,1,2,\ldots$, and multi-index set $\mathcal{A}\subset \mathbb{N}^{m}$, to extend the polynomial trend function for multivariable approximations, we can use tensor-product expansion that includes all combinations of the one-dimensional polynomial. All of the $\Psi_{i}(\boldsymbol{x})$ in Eq.~\ref{eq:UK} are multidimensional polynomials as the products of the one-dimensional polynomials, which can be constructed by using either monic or one-dimensional orthogonal polynomials. The multidimensional polynomials are then defined as
 	\begin{equation}
 		\Psi_{\boldsymbol{\zeta}}(\boldsymbol{x}) = \psi_{\zeta_{1}}^{(i)}(x_{1})\times\ldots\times\psi_{\zeta_{m}}^{(m)}(x_{m})
 	\end{equation}
 	for the one-dimensional Legendre polynomials, or 
 	\begin{equation}
 	\Psi_{\boldsymbol{\zeta}}(\boldsymbol{x}) = \phi_{\zeta_{1}}^{(i)}(x_{1})\times\ldots\times\phi_{\zeta_{m}}^{(m)}(x_{m})
 	\end{equation}
 	for the monic polynomials.
 	
 The polynomial terms can then be expanded by using the tensor-product operator of order $p$, that is
 	\begin{equation}
 	\mathcal{A}_{p} \equiv \{\boldsymbol{\zeta}\in  \mathbb{N}^{m}:\zeta_{j}\leq p, j = 1,\ldots,m\}.
 	\end{equation}
    Here, we use a fixed $p$ value for each dimension, although $p$ can differ for each dimension. Besides tensor-product expansion, total-order expansion that preserves the basis of polynomials up to a fixed total-order specification can be used as an alternative to the tensor-product operator. The index set for the total-order expansion of order $p$ is  defined as
 	\begin{equation}
 	\mathcal{A}_{p}\equiv \{\boldsymbol{\zeta}\in  \mathbb{N}^{m}:||\boldsymbol{\zeta}||\leq p\},
 	\end{equation}
 	where $||\boldsymbol{\zeta}||=\zeta_{1}+\ldots+\zeta_{m}$. 
 	
 	To further reduce the number of terms, a hyperbolic scheme~\citep{blatman2011adaptive} can be defined as
 	\begin{equation}
 	\mathcal{A}_{p,\nu}  \equiv \{\boldsymbol{\zeta}\in  \mathbb{N}^{m}:||\boldsymbol{\zeta}||_{\nu}\leq p\},
 	\end{equation}
 	where
 	\begin{equation}
 	||\boldsymbol{\zeta}||_{\nu}\equiv\left(\sum\limits_{i=1}^{m}\zeta_{i}^{\nu}\right)^{\frac{1}{\nu}},
 	\end{equation}
	and $\nu$ is a scalar in the range $(0,1]$. Here, the value of $\nu$ determines the number of polynomial terms to be retained. 
 	 	
 	The BK implementation available via the ooDACE toolbox limits the candidate trend function set to only two-factor interactions, although the value of $p$ can be set higher than two. Therefore, for example, the special cases considering linear effects, quadratic effects, and two-factor interactions would result in $2m^2$ candidate features (excluding the constant term), which is similar to the tensor-product expansion but with higher-factor interactions  eliminated from the candidate feature set. Since ooDACE implements this method to generate the polynomial trend function, in this paper, we also compare the performance of BK and PCK with this trend function generation method to provide a fair comparison between the two UK methods.
 	
 	In this paper, a trend function set with a defined value of $p$ indicates that the trend function set has a polynomial term with maximum order $p$, however, the cardinality of the candidate trend function set differs for each trend function generation method, even with the same value of $p$. In this regard, applying tensor-product, total-order expansion, or an expansion with maximum two-factor interactions will generate a trend function set with different cardinality for the same value of $p$.
 	
 	\subsubsection{Blind Kriging}
 	For the explanation of BK, we primarily refer to the works of~\cite{joseph2008blind} and~\cite{couckuyt2012blind}. BK employs automatic trend function selection from the candidate set via Bayesian forward selection, and then selects the best UK model that yields the lowest leave-one-out cross-validation (LOOCV) error. The BK model itself is particularly useful when the cardinality of the candidate trend function set exceeds the available sample size by providing only the relevant polynomial trend function set.  
 	
 Let us consider a trend function in the form of the following linear model
 	\begin{equation}
 	g(\boldsymbol{x}) = \alpha_{0}\Psi_{0}({\boldsymbol{x}})+\sum_{i=1}^{r}\alpha_{i}\Psi_{i}({\boldsymbol{x}})+\sum_{i=1}^{t}\beta_{i}b_{i}({\boldsymbol{x}}),
 	\end{equation}
 	
 	where $r+1$ is the size of existing trend functions, $\boldsymbol{b}(\boldsymbol{x})=\{b_{1}(\boldsymbol{x}),\ldots,b_{t}(\boldsymbol{x})\}$ is the set of candidate functions, and $\boldsymbol{\beta}= \{\beta_{1}(\boldsymbol{x}),\ldots,\beta_{t}(\boldsymbol{x}) \}^{T}$ is the vector of corresponding coefficients. Here, $\boldsymbol{\alpha}$ have already been determined, and the task is to select new terms to be included in the trend functions. In the explanation that follows, the candidate trend function considers only linear effects, quadratic effects, and two-factor interactions; however, it is also possible to build candidate sets with such higher-order effects and interactions. 

	As explained in~\cite{joseph2008blind}, the sample data is scaled to the interval $[1,3]$. The encoded samples for the linear and quadratic effects can then, respectively, be defined as
	 	\begin{equation}
	 	\begin{split}
	 	x_{jl} & = \frac{\sqrt{3}}{\sqrt{2}}(x_{j}-2), \\
	 	x_{jq} & = \frac{1}{\sqrt{2}}(3(x_{j}-2)^{2}-2). \\
	 	\end{split}
	 	\end{equation}
	Other terms, such as two-factor interaction terms, can be constructed as the products of these basic terms. To simultaneously estimate the $t$ effects from a Bayesian standpoint, a Gaussian prior distribution is then introduced for $\boldsymbol{\beta}(\boldsymbol{x})$,
	\begin{equation}
	\boldsymbol{\beta}\sim \mathcal{N}(\boldsymbol{0},\tau^{2}\boldsymbol{K}),
	\end{equation}
	where $\boldsymbol{K}$ is a $(t+1)\times(t+1)$ diagonal matrix. The construction of matrix $\boldsymbol{K}$ is defined below. First, the correlation function is assumed to has a product correlation structure of $r(\boldsymbol{h})=\prod_{i=1}^{m} r_{i}(h_{i})$. If we define $\boldsymbol{l}_{\textbf{i}}$ as the vector with element $l_{ij}=1$ if $\beta_{i}$ includes the linear effect of factor $j$ and 0 otherwise, and  $\boldsymbol{q}_{\textbf{i}}$ as the vector with element $q_{ij}=1$ if $\beta_{i}$ includes the quadratic effect of factor $j$ and 0 otherwise, the matrix 	$\boldsymbol{K}$ can then be defined as
		\begin{equation}
	 	\boldsymbol{K} = 
	 	\begin{bmatrix}
	 	\textbf{k}_{l}^{l_{1}} \cdot \textbf{k}_{q}^{q_{1}} & 0 & \ldots & 0 \\
	 	0 & \ddots & 0 & \vdots \\
	 	\vdots & 0 & \ddots & 0 \\
	 	0 & \ldots & 0 & 	\textbf{k}_{l}^{l_{t+1}}\cdot		\textbf{k}_{q}^{q_{t+1}} \\
	 	\end{bmatrix}.
	 	\end{equation}
where vectors $\textbf{k}_{l}$ and $\textbf{k}_{q}$ are respectively defined as	
 	\begin{equation}
 	\begin{split}
 	\textbf{k}_{l} & = \frac{3-3r(2)}{3+4r(1)+2r(2)},\\
 	\textbf{k}_{q} & = \frac{3-4r(1)+r(2)}{3+4r(1)+2r(2)}.\\
 	\end{split}
 	\end{equation}
From this, the posterior mean of $\boldsymbol{\beta}$ can then be estimated as
	 	\begin{equation}
	 	\begin{split}	
	 	\hat{\boldsymbol{\beta}} & = \frac{\tau^{2}}{\sigma^{2}} \boldsymbol{K} \boldsymbol{M}_{c}' \boldsymbol{R}^{-1}(\boldsymbol{y}-\boldsymbol{M}\boldsymbol{\alpha}), \\ 	
	 	var(\hat{\boldsymbol{\beta}}) & = \tau^{2} \big(\boldsymbol{K}- \frac{\tau^{2}}{\sigma^{2}} \boldsymbol{K} \boldsymbol{M}_{c}'\boldsymbol{R}^{-1}\boldsymbol{M}_{c}\boldsymbol{K} \big),\\	
	 	\end{split}
	 	\end{equation} 	
 where $\boldsymbol{M}_{c}$ is the model matrix of all candidate terms and $\boldsymbol{M}$ is the model matrix of all terms in the existing trend function of the ED.
 
 The output of this procedure is the set of coefficients $\boldsymbol{\beta}$, which denotes the importance of the associated trend function given the current experimental design. Based on $\boldsymbol{\beta}$, we can extract the most promising feature to be included in the trend functions. The BK algorithm can then be summarized as follows:
 	
 	\begin{enumerate}
 		\item{Build an initial design of experiments $\boldsymbol{\mathcal{X}}$ and $\boldsymbol{y}$.}
 		\item{Perform Bayesian forward selection using the candidate set $\mathcal{A}$.}
 		\item{Build a new UK model at each iteration of the Bayesian forward selection algorithm with the current polynomial trend function set.}
 		\item{Compute the LOOCV error for each UK surrogate model.}
 		\item{Select the UK surrogate model that has the lowest LOOCV error as the final surrogate model.}
 	\end{enumerate}    	
 	
 	To ensure optimal implementation of BK, we used several value of $p$ and then selected the optimal value of $p$, i.e., the one with the lowest LOOCV error. Further, we used the ooDACE toolbox~\citep{couckuyt2014oodace} to create the BK surrogate model. Again, note that the ooDACE toolbox creates the candidate trend function set with only the maximum two-factor interactions included. 
 	
 	\subsubsection{Polynomial-Chaos-Kriging}
 	In our explanation of PCK, we primarily refer to the works of~\cite{schobi2015polynomial} and~\cite{kersaudy2015new}. The use of an orthogonal polynomial trend function for UK was first proposed by~\cite{schobi2015polynomial}. The type of orthogonal polynomials used in PCK depends on the assigned input probability distribution. When we want to use PCK for an optimization problem, Legendre polynomials are the most appropriate due to the bounded and uniform search domain of the optimization problem. Here, the PCK can be built using one of two approaches, that is, either sequential PCK or optimal PCK. The simplest and fastest approach is sequential PCK, which directly uses the trend function returned by pure LARS-PCE~\citep{blatman2011adaptive}. The sequential approach assumes that the set of trend functions returned by pure PCE approximation and LARS is also the optimal set for the PCK surrogate model. An optimal but more expensive approach is to build a new PCK at each iteration of the LARS algorithm. In this paper, we use the optimal PCK to ensure a high-quality PCK surrogate model at each EGO iteration.
 		
 	An important part of obtaining the trend function set $\boldsymbol{\Psi}(\boldsymbol{x})$ for PCK is to make use of the LARS algorithm. LARS is especially useful when the dimensionality of a problem is high since LARS works by identifying a set of the most influential terms to be incorporated into the regression scheme~\citep{blatman2011adaptive}. Here, one must first prepare an a priori set of polynomial terms, and the LARS algorithm automatically selects the subset of terms that yields the lowest leave-one-out error. Also, note that LARS is especially useful for our application that involves EGO in high dimensions, which we describe later. For a specified prediction, the LARS algorithm is summarized as follows:
 		
 	\begin{enumerate}
 		\item Build the candidate polynomial set $\mathcal{A}$ of degree $p$.
 		\item Set coefficients $\alpha_{0},...,\alpha_{P-1}$ to 0 and initial residual vector $\boldsymbol{\epsilon}^{(0)} = \boldsymbol{y}$.
 		\item Select polynomial $\Psi_{\boldsymbol{\zeta}^{(j_{1})}}$ that is  most correlated with $\boldsymbol{y}$.
 		\item Move coefficient $\alpha_{\boldsymbol{\zeta}^{(j_{1})}}$ in the direction of predictor $\Psi_{\boldsymbol{\zeta}^{(j_{1})}}$ until the current residual $\boldsymbol{\epsilon}^{(1)}=\boldsymbol{y}-\hat{\alpha}_{\boldsymbol{\zeta}^{(j_{1})}}\Psi_{\boldsymbol{\zeta}^{(j_{1})}}(\boldsymbol{\mathcal{X}})$ is perfectly correlated with $\Psi_{\boldsymbol{\zeta}^{(j_{1})}}$ and another predictor $\Psi_{\boldsymbol{\zeta}^{(j_{2})}}$.
 		\item Move jointly $\{\alpha_{\boldsymbol{\zeta}^{(j_{1})}},\alpha_{\boldsymbol{\zeta}^{(j_{2})}} \}^{T}$ in the direction  defined by their joint least-squares coefficient until some other predictor $\Psi_{\boldsymbol{\zeta}^{(j_{3})}}$ has high correlation with the current residual.
 		\item Repeat steps 3-5 above until $\text{min}(P,n-1)$ is achieved.
 	\end{enumerate}
 	
 	The LARS algorithm is combined with UK to create an optimal UK surrogate model based on the orthogonal polynomial~\citep{schobi2015polynomial, kersaudy2015new}. With this method, the trend function coefficients for each possible term are calculated using GLS given the selected polynomials. The difference between PCK and LARS-PCE is that a new UK is built at each iteration of the LARS algorithm instead of the pure PCE approximation. Using this approach, the process of building the polynomial set takes the correlation of residuals given by the kernels into account. The algorithm for building a PCK surrogate model can then be defined as follows:
 	\begin{enumerate}
 	 		\item{Build the initial design of experiments $\boldsymbol{\mathcal{X}}$ and $\boldsymbol{y}$.}
 	 		\item{Perform the LARS algorithm using defined candidate set $\mathcal{A}$}.
 	 		\item{Build a new PCK model at each iteration of the LARS algorithm with the current polynomial set.}
 	 		\item{Compute the LOOCV error for each PCK surrogate model.}
 	 		\item{Select the PCK surrogate model with the lowest LOOCV error as the final surrogate model.}
 	\end{enumerate}
 	Similar to the pure LARS-PCE approach, we tested different values of $p$ to identify the best possible combination of polynomial sets for PCK. Here, we used either: the tensor product, total order expansion, or the expansion with maximum two-factor interactions to build the candidate polynomial set. In our research, we developed our own PCK code by modifying the OK code detailed by~\cite{forrester2008engineering} as a basis.
 	
 	BK and optimal PCK can improve the quality of the UK surrogate model, though we have the additional computational cost of building the UK model itself. The high computational cost primarily stems from the need to train the hyperparameters at each iteration of the Bayesian forward selection or LARS algorithm. The computational cost significantly increases if several values of $p$ are tested to further discover the polynomial set with the lowest LOOCV error. It is worth noting that in real-world applications, the computational cost to construct UK with automatic trend function selection can be considered negligible when compared to the simulation cost of high-fidelity models. However, acceleration of hyperparameters training benefits our study, since we need to repeat the experiment multiple times. Therefore, it is necessary to accelerate this process without sacrificing the accuracy of the UK surrogate model. To deal with this issue, we developed a simplified GA+BFGS strategy to select the trend function and optimize the hyperparameters for PCK. The simplified GA+BFGS strategy employs a GA on the first iteration, while applying BFGS on subsequent iterations using the optimum hyperparameters from each previous iteration as its initial solution. This approach led to the acceleration of the construction process for PCK while returning a similar LOOCV error as compared to the exhaustive GA+BFGS strategy that employs GA+BFGS at every iteration of the BK/PCK algorithm. However, we only applied our proposed hyperparameter optimization strategy to the PCK model since we used the third-party ooDACE code to build the BK surrogate model. We used the exhaustive GA+local search strategy of ooDACE to train the hyperparameters of OK to ensure the best quality BK surrogate model for our work. The details of the simplified GA+BFGS strategy and results from numerical experiment are explained in detail in Appendix A.

 	\section{Universal Kriging for efficient global optimization}
 	\label{sec:3}
 	\subsection{Framework}
 	The primary contribution of this paper is to investigate the capabilities of UK in EGO to solve single-objective optimization problems given the constraint of a limited budget. More specifically, we studied the implementation of UK with a fixed predefined polynomial set (first- and second-order polynomial built using total-order expansion) and the methods with automatic trend function selection incorporated into the EGO framework. Here, UK with a fixed predefined polynomial set uses the Legendre polynomials. On the other hand, BK and PCK employ multidimensional polynomials from monic and Legendre polynomials, respectively. We investigated two types of UK with automatic trend function selection, that is, BK and PCK. For convenience, we refer to our UK-EGO implementation that employ BK and PCK as BK-EGO and PCK-EGO, respectively. Since the key difference between standard EGO and these two models lies only in the type of Kriging model employed, we refer to the original paper on EGO for the main algorithm of the optimizer~\citep{jones1998efficient}. All EGO variants explained in our paper use the EI as the metric to be optimized, that is
 	\begin{equation}
 	E[I(\boldsymbol{x})] = (y_{min}-\hat{f}(\boldsymbol{x})) \Theta \bigg(\frac{y_{min}-\hat{f}(\boldsymbol{x})}{\hat{s}(\boldsymbol{x})} \bigg) + \hat{s}(\boldsymbol{x})\varphi \bigg(\frac{y_{min}-\hat{f}(\boldsymbol{x})}{\hat{s}(\boldsymbol{x})} \bigg),
 	\end{equation}	
 	where $y_{min}$ is the best solution identified so far, and $\Theta(.)$ and $\varphi(.)$ are the cumulative distribution and probability density functions of the standard normal distribution, respectively. Here, EI can be evaluated by using the error function as
 	\begin{multline}
 	E[I(\boldsymbol{x})] = (y_{min}-\hat{f}(\boldsymbol{x}))\bigg[\frac{1}{2}+\frac{1}{2}\text{erf}\bigg(\frac{y_{min}-\hat{f}(\boldsymbol{x})} {\hat{s}(\boldsymbol{x})\sqrt{2}} \bigg)\bigg]\\
 	+\hat{s}(\boldsymbol{x})\frac{1}{\sqrt{2\pi}} \text{exp}\bigg[\frac{-(y_{min}-\hat{f}(\boldsymbol{x}))^{2}}{2\hat{s}(\boldsymbol{x})^{2}} \bigg].
 	\end{multline}
	The next sample to be added to the ED is then found by maximizing the EI, that is,
 	\begin{equation}
 	\underset{\boldsymbol{x}}{\mbox{arg max }} E[I(\boldsymbol{x})].
 	\end{equation}
 	
 	Optimization using the EI metric takes advantage of the prediction and mean-squared error of the Kriging model when searching for the optimum point of the surrogate model. EI-based optimization ensures a balance between exploration and exploitation of Kriging based search. Therefore, EI-based optimization has a higher likelihood to escape from local or false optima versus that of simple prediction-based search; however, if either BK or PCK is used as a surrogate model, the algorithm must be modified slightly. Note that since the difference between all EGO variants considered in our paper hinges on the choice of surrogate model and not on the specific criterion to be optimized, other criteria such as entropy improvement or bootstrapped EI, are also directly applicable to the UK surrogate model. Nonetheless, we suggest that more studies be conducted to further explore when different criteria are used with the PCK surrogate model.
 	
 	EGO works by first preparing initial ED $\boldsymbol{\mathcal{X}}$ and $\boldsymbol{y}$. A Kriging surrogate model (i.e., UK or OK) is then built using this initial dataset. After the Kriging surrogate model has been built, the solution with the maximum EI value is then searched using a global and/or local optimizer. This new solution is then evaluated and added to the ED. At each EGO iteration, the minimum objective value and corresponding solution are recorded. This process is then repeated until the computational budget is exhausted or no further change is observed in the optimum value. 
 	
 	The overall algorithm of EGO with UK is similar to standard EGO with OK with the key difference being the surrogate model used. EGO with UK of first- and second-order polynomials is relatively straightforward to perform, that is, one just constantly employs UK with either the first- or the second-order polynomial at each EGO iteration. Both BK-EGO and PCK-EGO have an additional step in which UK is built through an automatic trend function selection procedure before a new solution is added. The final polynomial set for BK and PCK is found by choosing the polynomial set with the lowest LOOCV error $e_{LOO}$ from various values of $p$, where this polynomial set is denoted as $\mathcal{A}_{p_{min}}$. Next, the BK or PCK surrogate model with the lowest LOOCV error is selected. This chosen surrogate model is then searched by maximizing $EI$ to identify the next solution to be added. This step is then repeated until the computational budget is exhausted. From the above, BK/PCK-EGO algorithm is summarized as Algorithm~\ref{alg:PCKEGO}.
 	
 	\begin{algorithm}
 		Start the iteration counter $o = 1$;\\
 		Build the initial ED $\boldsymbol{\mathcal{X}}^{(o)}$;\\
 		Evaluate the output $\boldsymbol{y}^{(o)}$;\\
 		Find the best solution $\boldsymbol{x}_{best}$  among $\boldsymbol{\mathcal{X}}^{(o)}$, and set $y_{min}=f(\boldsymbol{x}_{best})$ \\
 		
 		\While{computational budget is not exhausted}{
 			
 			\textbf{/****Universal Kriging phase****/};\\
 			\For{$p \gets 0 \textrm{ to } p_{max} $}{
 				Build the candidate polynomial set $\mathcal{A}_{p}$;\\
 				Set the initial trend function $\boldsymbol{\Psi}=\{\Psi_{0}\}$ (OK);\\
 				\For{$i \gets 0 \textrm{ to } P-1$}{
 					Build a Kriging surrogate model $\hat{y}^{(i)}(\boldsymbol{x})$ using $\boldsymbol{\mathcal{X}}^{(o)}$,$\boldsymbol{y}^{(o)}$, and trend function $\{\Psi_{0},\ldots,\Psi_{i} \}$;\\
 					Calculate the $e_{LOO,p}(i)$ from $\hat{y}^{(i)}(\boldsymbol{x})$;\\
 					Choose the next trend function $\Psi_{i+1}$ to be added into the current set $\boldsymbol{\Psi}$ using LARS or Bayesian forward selection.\\
 					Store $e_{LOO,p,min} \equiv \text{min}(e_{LOO,p})$;\\
 					\If {$e_{LOO,p,min}\equiv e_{LOO,p}(i)$}{Set $i_{p,min}=i$;\\
 					}
 					\If {$e_{LOO,p}$ increases thrice in a row}{
 						\textbf{Break for loop}}
 				}
 				Set the optimum UK surrogate model of order $p$ with the trend function $\{\Psi_{0},\ldots,\Psi_{i_{p,min}} \}$
 			}
 			
 			Choose the UK surrogate model with the lowest $e_{LOO}$ from various values of $p$;
 			Optimize the EI metric using the selected PCK surrogate model to find the next sample $\boldsymbol{x}^{(o+1)}$ to be added to the ED;\\
 			Evaluate $\boldsymbol{x}^{(o+1)}$;\\
 			if $f(\boldsymbol{x}^{(o+1)}) < f(\boldsymbol{x}_{best})$, set $\boldsymbol{x}_{best} = \boldsymbol{x}^{(o+1)} $ and $y_{min}=f(\boldsymbol{x}^{(o+1)})$;\\
 			Add the new sample to the ED $\boldsymbol{\mathcal{X}}^{(o+1)} = \boldsymbol{\mathcal{X}}^{(o)}  \cup \boldsymbol{x}^{(o+1)} $;\\
 			Increment the iteration counter $o=o+1$;\\
 		}
 		Return the best solution $\boldsymbol{x}_{best}$ and its output $y_{min}$.
 		\caption{UK-EGO with automatic trend function selection (PCK/BK) main loop.}
 		\label{alg:PCKEGO}
 	\end{algorithm}
 	
 	In our paper, we optimize the EI metric using a hybrid combination of GA and BFGS, where BFGS uses the final solution found by the GA as its initial solution. The key advantage of using the BK-EGO and PCK-EGO method is that each surrogate model is a Kriging model, meaning that each surrogate model has an uncertainty structure that can be employed for EI-based optimization. 
 	
 	The LOOCV method assists in the process of selecting the best surrogate model for the BK-EGO and PCK-EGO that hopefully yields the lowest actual error. One key advantage of using LOOCV error is that it can be analytically calculated within a Kriging framework, as described in detail by~\cite{dubrule1983cross}. The LOOCV error can then be directly plugged into any error metric, where in this paper we opt for the root mean squared error (RMSE) as the error metric, that is,
 	 	\begin{equation}
 	 	e_{LOO}= \sqrt{\frac{1}{n}\sum_{i=1}^{n} (f(\boldsymbol{x}^{(i)})-\hat{f}^{(-i)}(\boldsymbol{x}^{(i)}))^{2}},
 	 	\end{equation}
 	where $\hat{f}^{(-i)}$ is the Kriging prediction with the sample $i$ is removed from the original ED.
 	
 	\section{Optimization performance analysis on test problems}
 	\label{sec:4}
 	As the core of our work, we studied the performance of various UK schemes to optimize several synthetic test functions and aerodynamic problem within the EGO framework. Experiments on synthetic functions is necessary since they are cheap to evaluate, which allows us to perform numerous independent runs so the results can be analyzed statistically. On the other hand, studies focused on real-world problem are necessary to further assess the performance of EGO with UK. In this paper, the real-world problem considered is the aerodynamic efficiency optimization of a transonic airfoil. 
 	
 	Mathematical expressions for the given synthetic test problems are detailed in Appendix B, while the variables domain, initial sample size $N_{int}$, number of updates $N_{upd}$, and true global optimum values are shown in Table~\ref{table:resultsstatistics}. The first function is the Branin function, which has three global optima, and is relatively easy to solve via EGO in spite of its relatively complex trend. The second problem is the Sasena function, which exhibits a complex trend with one global optimum and three local optima. The third function is the Hosaki function, which features a highly nonlinear trend with one local optimum and one global optimum. Next, the Hartman-6 function is a challenging problem given the difficult location of its optimum solution. In this paper, we use a log-type transformation $y\mapsto -\text{log}(-y)$ for the Hartman-6 function. Finally, the last artificial test function is the minimization of the eight-dimensional borehole problem. For the Hosaki and borehole problems, we set the initial sample size lower than $10\times m$ rule to make the problem more difficult, since we cannot observe  clear performance differences between all algorithms using the $10\times m$ rule on these two functions. In general, we set $N_{upd}$ to 10 except for the Sasena and Hartman-6 problems since it takes longer  to converge to the near-optimum location on these two functions.
 	
 	For the PCK-EGO, we implemented two methods to generate the candidate polynomial sets for the higher-dimensional problems. The first method here is the total-order expansion as it is used in the original PCK algorithm. The second method uses the same candidate generation method as suggested in the original implementation of BK in the ooDACE toolbox, which limits the trend function up to the two-factor interaction. A fairer comparison of PCK-EGO and BK-EGO can be achieved if the initial candidate trend function is the same, which is why we implemented these two methods to generate candidate trend function sets for PCK.
 	
 	\begin{table}[h]
 		\centering
 		\caption{Specific test problems considered in our study.}
 		\label{table:resultsstatistics}
 		\begin{tabular}{llllll} 
 			\hline\noalign{\smallskip} 
 			No.  & Problem & Variables domain & $N_{int}$ & $N_{upd}$ & Optimum value \\ \hline\noalign{\smallskip}  \hline\noalign{\smallskip} 
 			1 &  Branin  & $[0,1]^{2}$ & 20 & 10 & 0.39788 \\
 			2 & Sasena  & $[0,5]^{2}$ & 20 & 20 & -1.4565 \\ 
 			3 & Hosaki  & $[0,5]^{2}$ & 12 & 10 & -2.3458 \\ 
 			4 & Hartman-6  & $[0,1]^{6}$ & 60 & 25 & -3.32237 \\
 			5 & Borehole (min.)  & See appendix B & 40 & 10 & 7.8198 \\
 			\hline\noalign{\smallskip} 
 		\end{tabular}
 		
 	\end{table}
 	
 	For the two-dimensional problems, we utilized tensor-product expansion with $p_{max}=4$ to construct the candidate polynomial set for PCK and BK; note that the candidate trend function size was the same for BK and PCK since the given problems are two-dimensional. Conversely, the candidate polynomial set of PCK for higher-dimensional problems was constructed using total-order expansion and the maximum two-factor interactions, with $p_{max}=3$ and $p_{max}=2$ for the Hartman-6 and borehole problem, respectively. For the ooDACE implementation of BK, we used the tensor product with maximum two-factor interactions for high-dimensional problem; hence we used it as is. To properly take the stochastic nature of the EGO-based algorithm into account, the results shown in this paper were obtained by averaging the results from 20 different runs, each with a different set of latin hypercube sampling (LHS) samples. These 20 different sets are identical for all EGO variants to ensure that we make a fair comparison between all algorithms. 
 	
 	We monitored the performance of the optimizer by using the improvement metric defined as
 	\begin{equation}
 	I = \frac{\big|f(\boldsymbol{x}_{opt})-f(\boldsymbol{x}_{best})\big|}{\big|f(\boldsymbol{x}_{opt})\big|},
 	\end{equation}	
 	where $f(\boldsymbol{x}_{opt})$ and $f(\boldsymbol{x}_{best})$ represent the true optimum solution of a given problem and the current best optimum. Here, the lower the value of $I$, the better the performance of the optimizer.
 	
 	Besides analyzing the optimization performance, we also analyze the approximation quality of OK and UK with the initial sample set. Our goal here is to investigate whether the approximation quality of the initial sample set has a clear relationship with the performance of the EGO. The quality of our surrogate model is measured using RMSE calculated as
 	 	\begin{equation}
 	 	RMSE = \sqrt{\frac{1}{n_{v}}\sum_{i=1}^{n_{v}} \big(f(\boldsymbol{x}^{(i)})-\hat{f}(\boldsymbol{x}^{(i)})\big)^{2}},
 	 	\end{equation} 	 	
 	where $n_{v}$ is the number of validation samples, which we fixed at 100.000.

 	\subsection{Studies in synthetic test problems}
\begin{figure}
 		\centering
	\begin{subfigure}{.495\columnwidth}
 			\includegraphics[width=1\columnwidth]{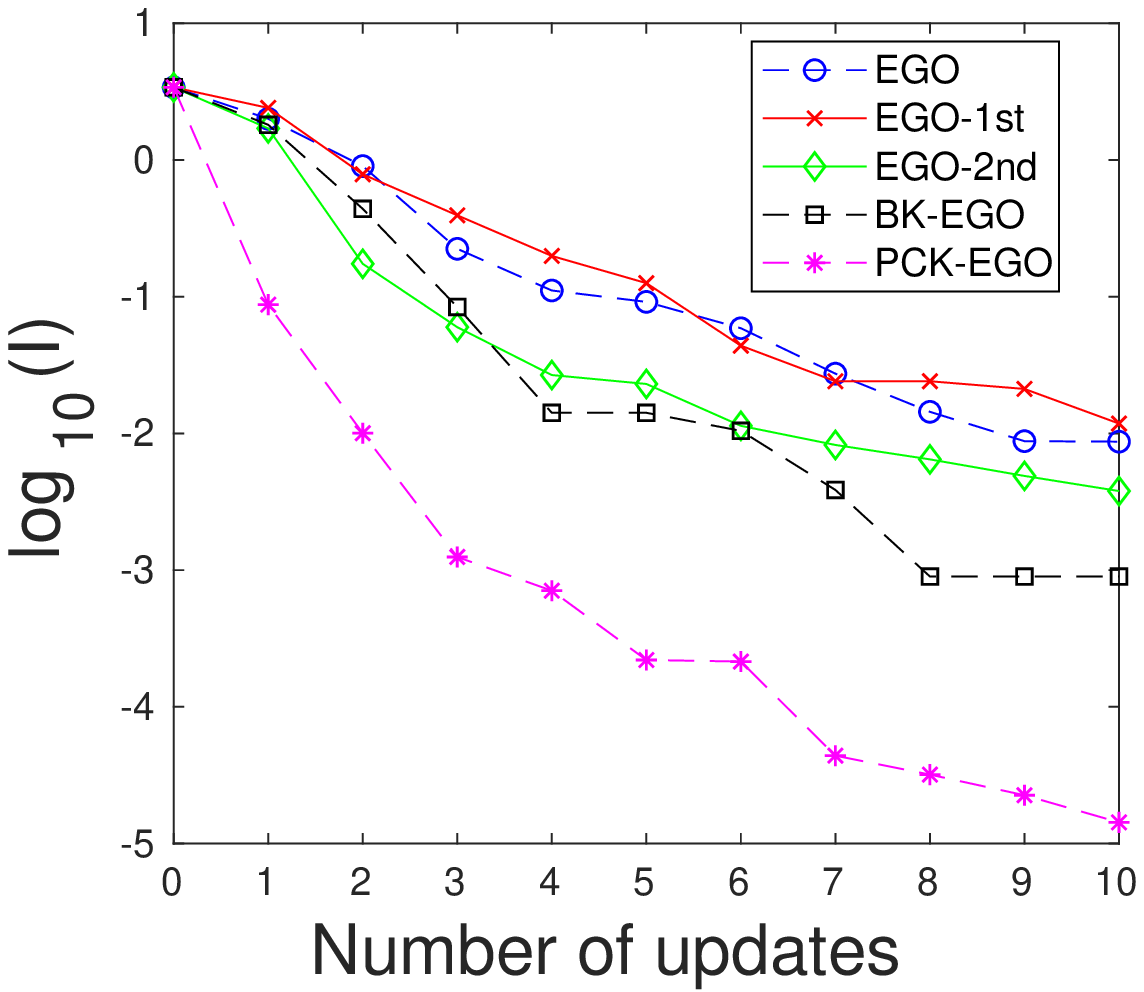}%
 			\caption{}%
 			\label{fig:BRANIN_UK_COMPAR}%
 		\end{subfigure}\hfill%
	\begin{subfigure}{.495\columnwidth}
 			\includegraphics[width=1.\columnwidth]{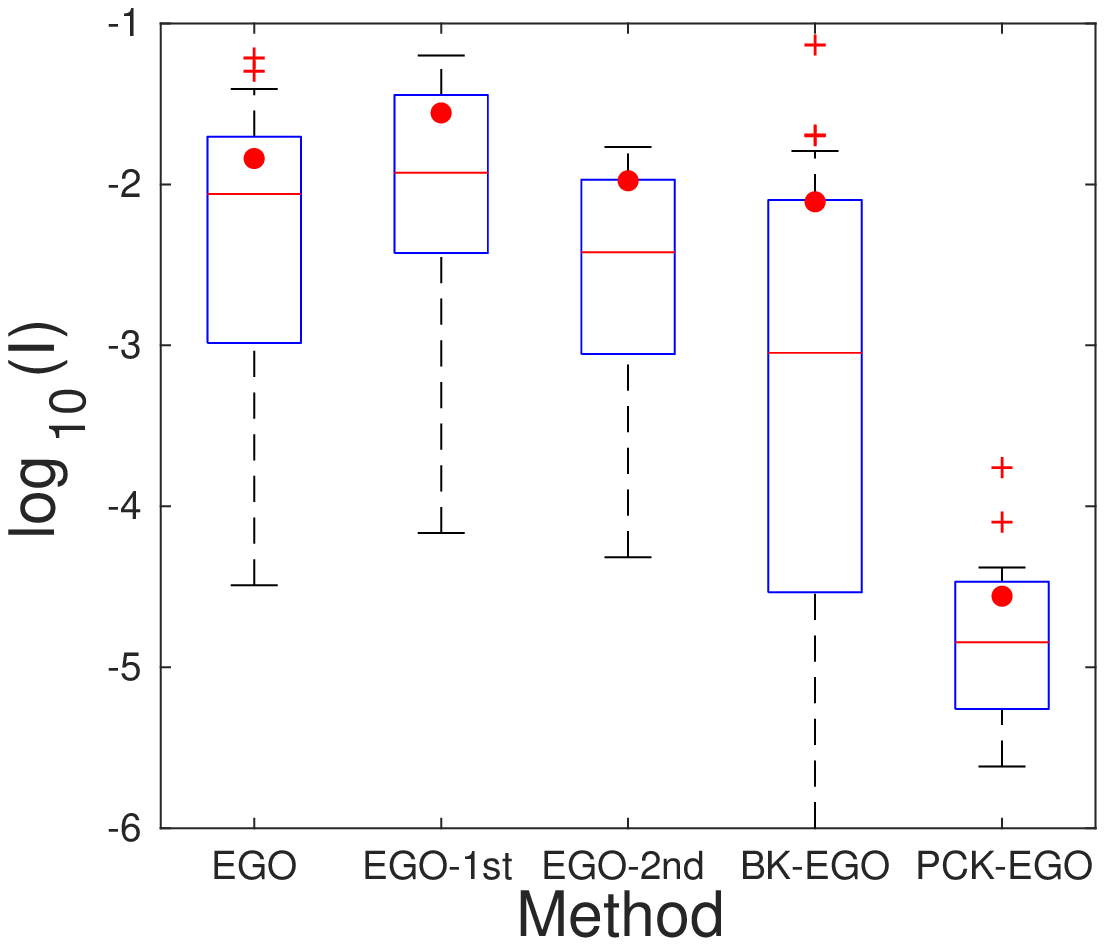}%
 			\caption{}%
 			\label{fig:BRANIN_UK_COMPAR_BOX}%
 		\end{subfigure}\hfill%
 			\begin{subfigure}{.49\columnwidth}
 		 			\includegraphics[width=1\columnwidth]{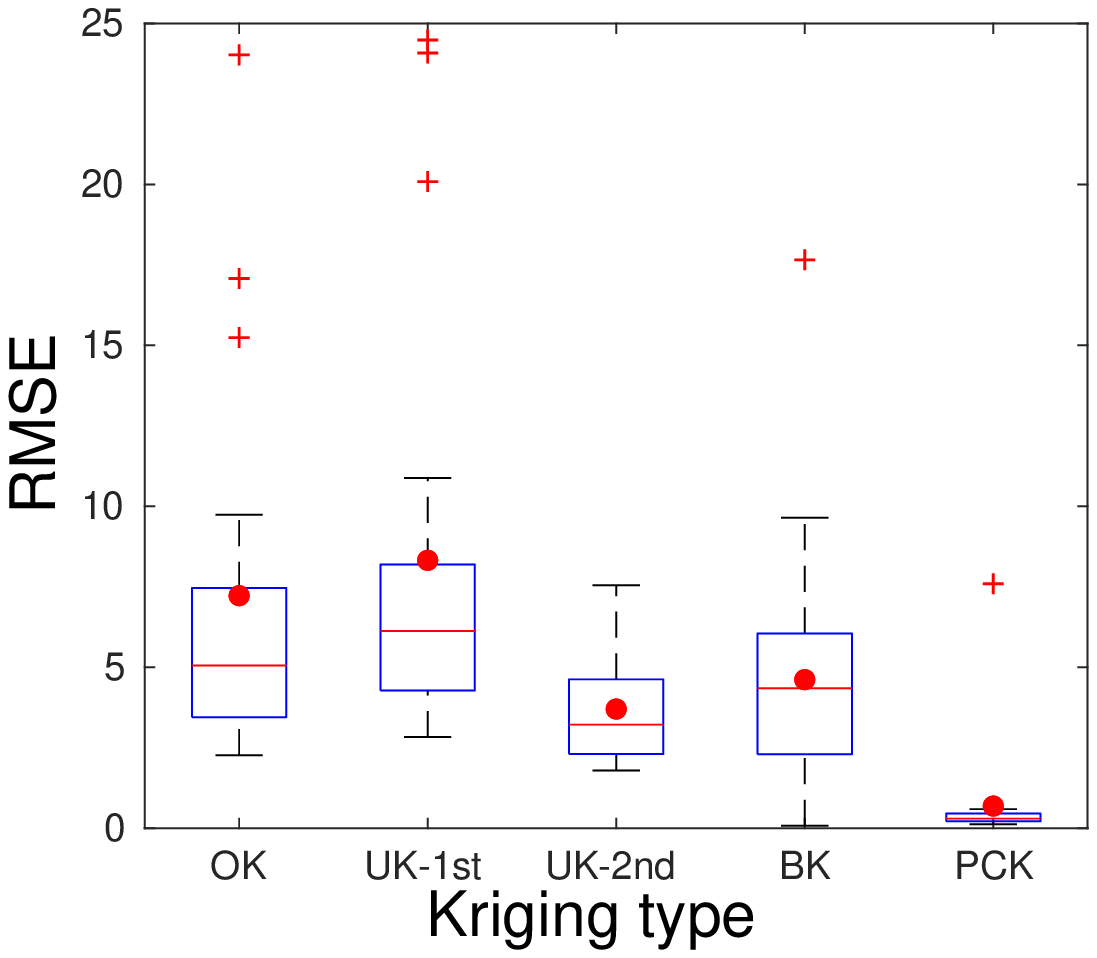}%
 		 			\caption{}%
 		 			\label{fig:BRANIN_ERROR_PLOT_INT}%
 		 	 	 			
 		 		\end{subfigure}\hfill%

 		\caption{Results obtained from various EGO schemes for the Branin function: \textbf{a} Convergence of the median, \textbf{b} boxplot, and \textbf{c} RMSE from the initial surrogate models.}
 		\label{fig:braninplot}
 	\end{figure}

\begin{figure}
 		\centering
 		\begin{subfigure}{.5\columnwidth}
 			\includegraphics[width=1\columnwidth]{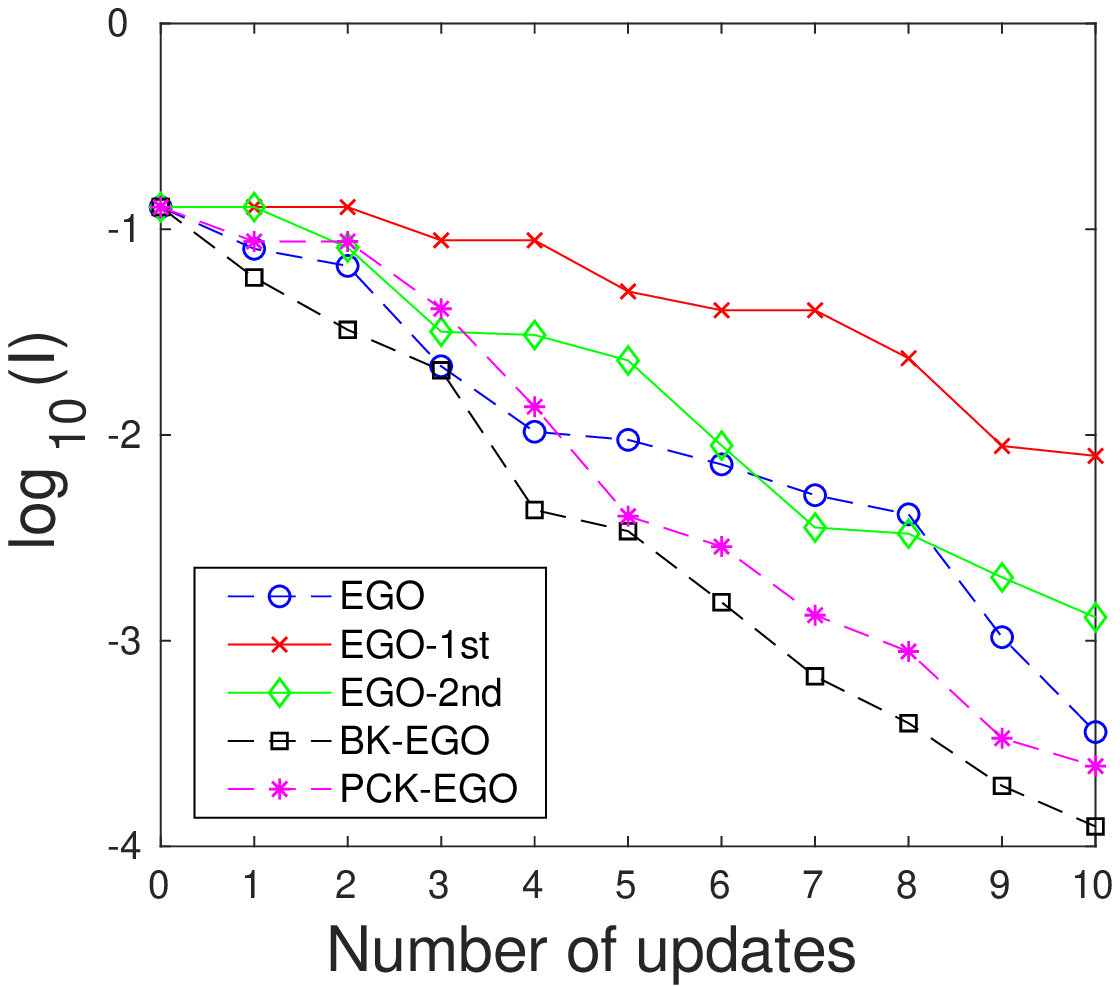}%
 			\caption{}%
 			\label{fig:HOSAKI_UK_COMPAR}%
 		\end{subfigure}\hfill%
 		\begin{subfigure}{.5\columnwidth}
 			\includegraphics[width=1\columnwidth]{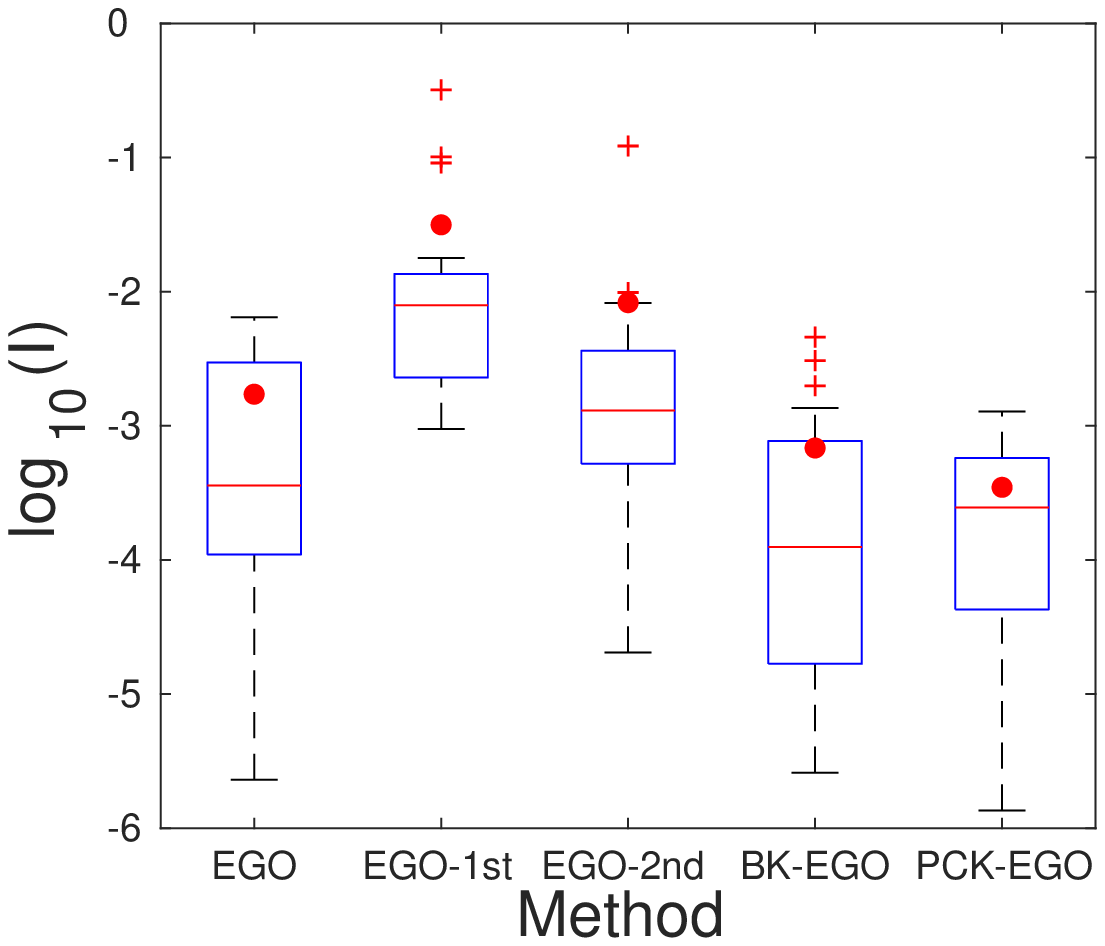}%
 			\caption{}%
 			\label{fig:HOSAKI_UK_COMPAR_BOX}%
 		\end{subfigure}\hfill%
	\begin{subfigure}{.5\columnwidth}
 			\includegraphics[width=1\columnwidth]{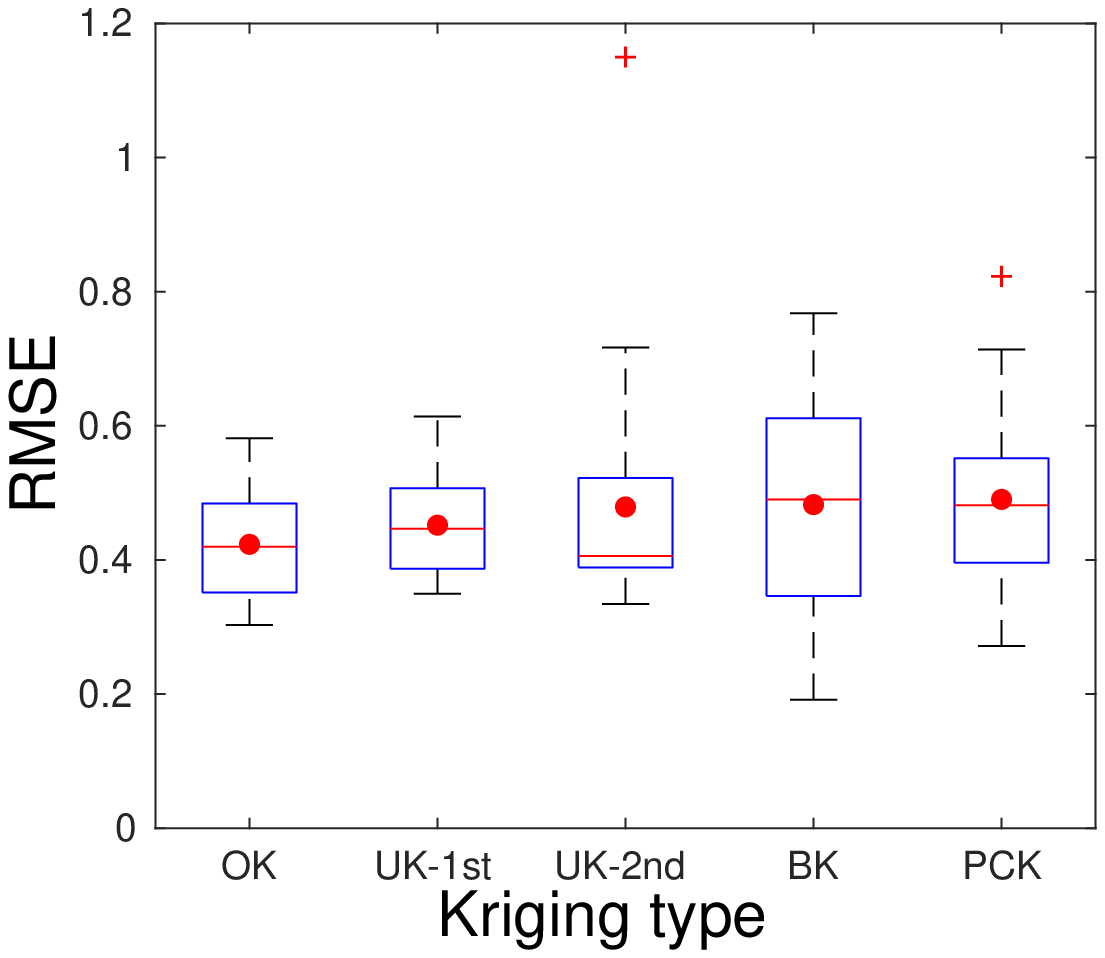}%
 			\caption{}%
 			\label{fig:HOSAKI_ERROR_PLOT_INT}%
 	 	 			
 		\end{subfigure}\hfill%
 		\caption{Results obtained from various EGO schemes for the Hosaki function: \textbf{a} Convergence of the median, \textbf{b} boxplot, and \textbf{c} RMSE from the initial surrogate models.}
 		\label{fig:hosakiplot}
 	\end{figure}

\begin{figure}
 		\centering
 		\begin{subfigure}{.5\columnwidth}
 			\includegraphics[width=1\columnwidth]{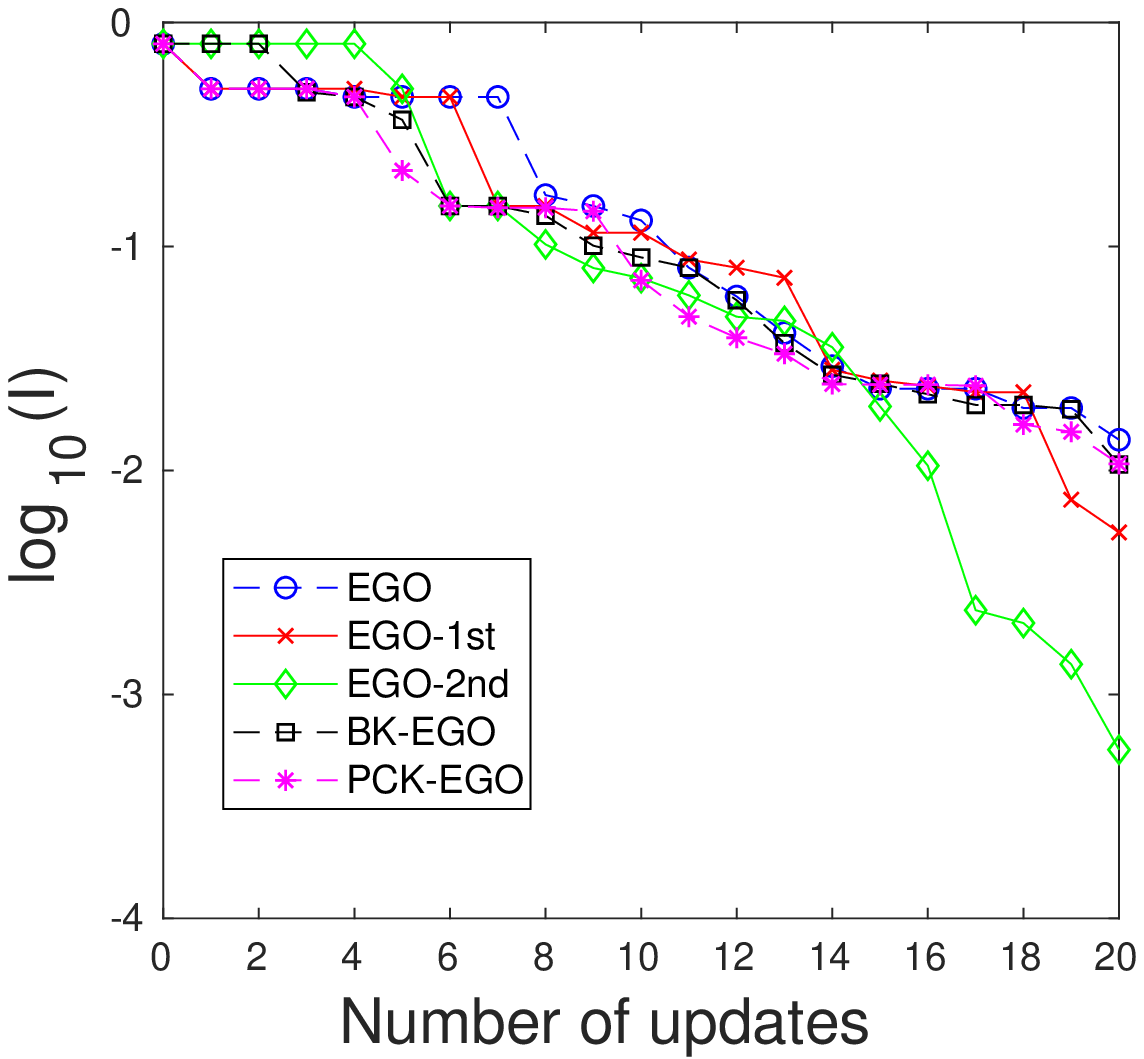}%
 			\caption{}%
 			\label{fig:SASENA_UK_COMPAR}%
 		\end{subfigure}\hfill%
 		\begin{subfigure}{.5\columnwidth}
 			\includegraphics[width=1\columnwidth]{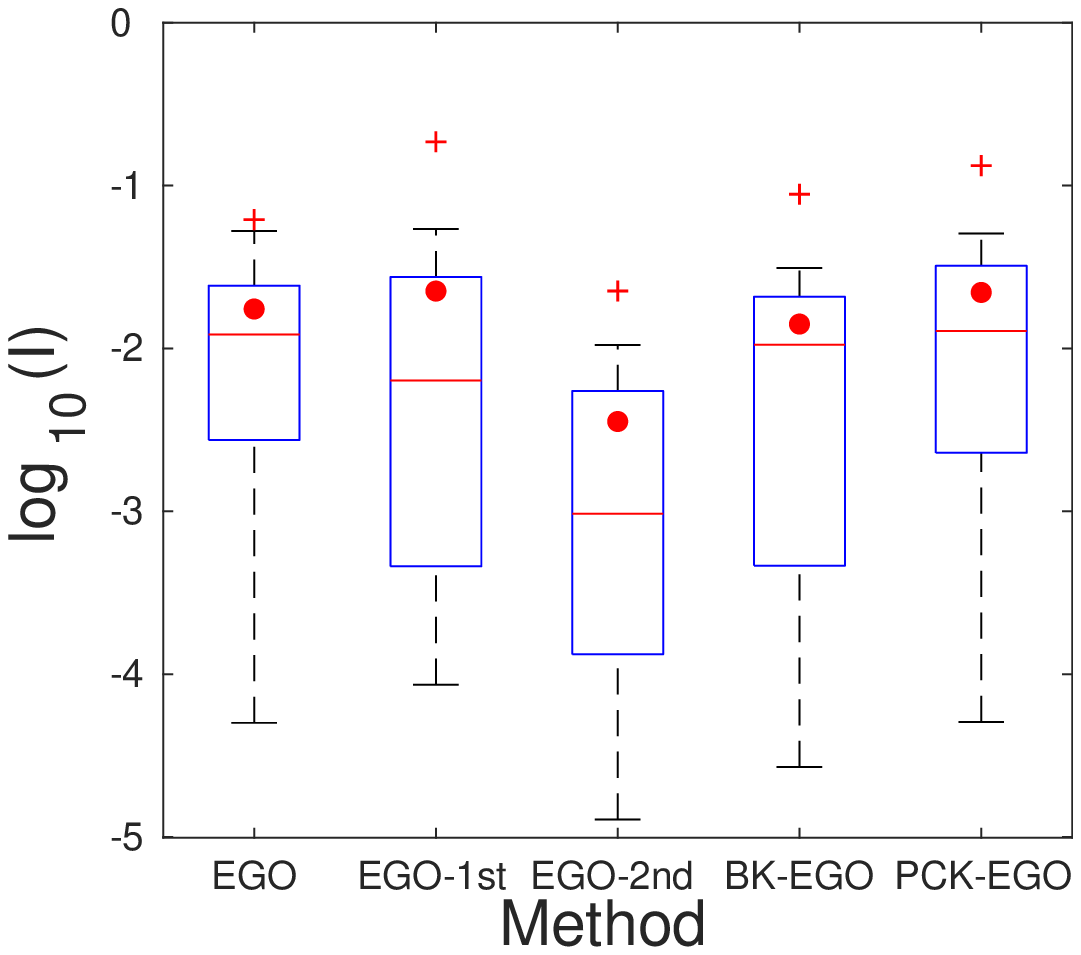}%
 			\caption{}%
 			\label{fig:SASENA_UK_COMPAR_BOX}%
 		\end{subfigure}\hfill%
	\begin{subfigure}{.5\columnwidth}
 			\includegraphics[width=1\columnwidth]{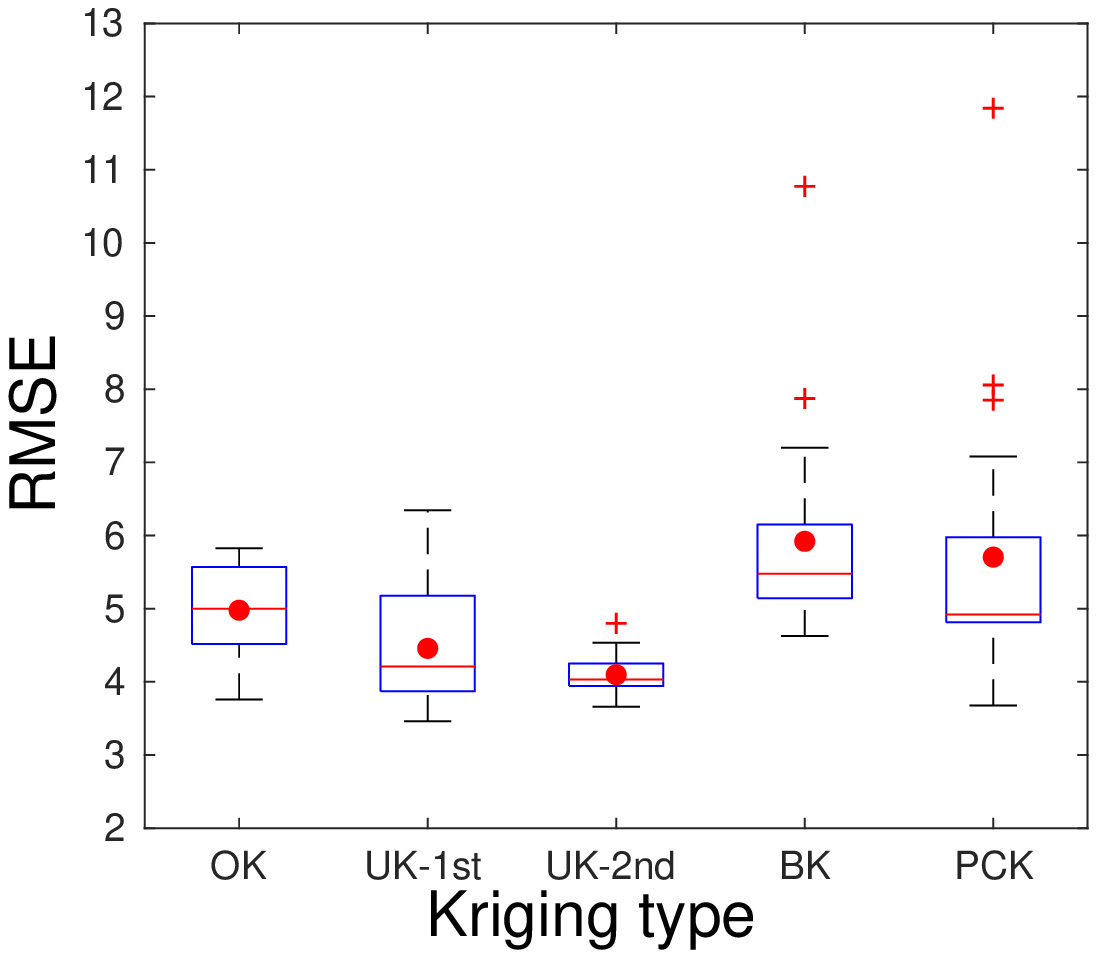}%
 			\caption{}%
 			\label{fig:SASENA_ERROR_PLOT_INT}%
 		\end{subfigure}\hfill%
 		\caption{Results obtained from various EGO schemes for the Sasena function: \textbf{a} Convergence of the median, \textbf{b} boxplot, and \textbf{c} RMSE from the initial surrogate models.}
 		\label{fig:sasenaplot}
 	\end{figure}

\begin{figure}
 		\centering
 		\begin{subfigure}{.5\columnwidth}
 			\includegraphics[width=1\columnwidth]{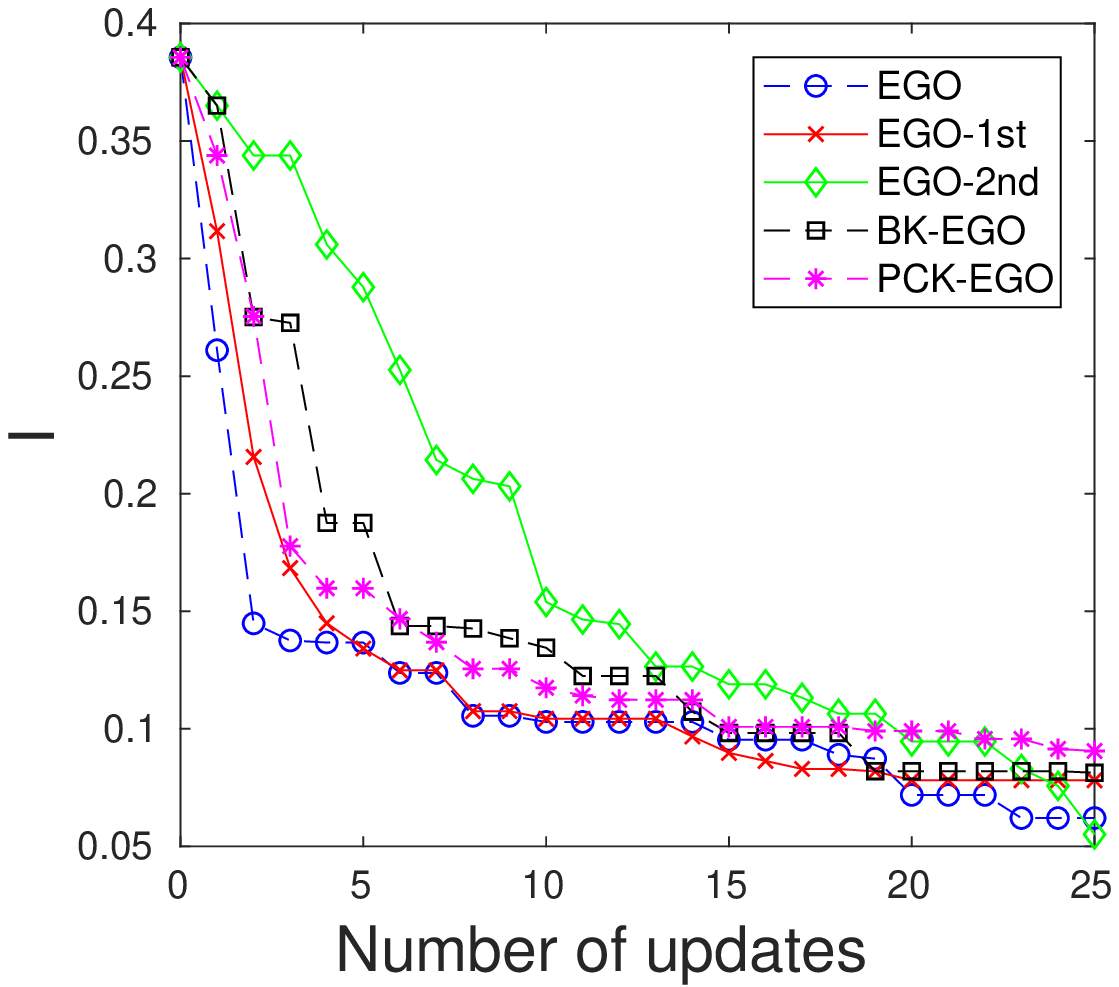}%
 			\caption{}%
 			\label{fig:HART_UK_COMPAR}%
 		\end{subfigure}\hfill%
 		\begin{subfigure}{.52\columnwidth}
 			\includegraphics[width=1\columnwidth]{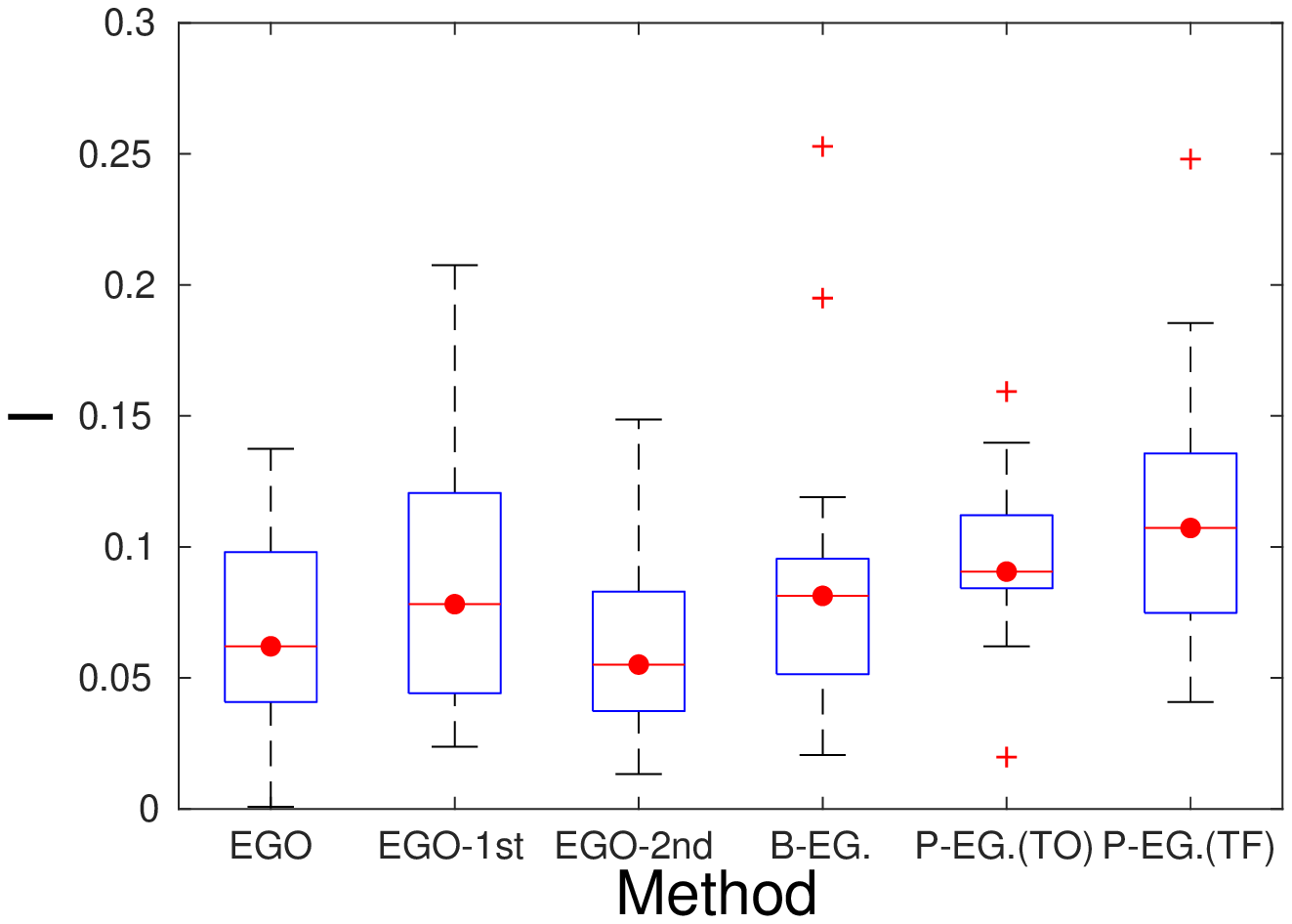}%
 			\caption{}%
 			\label{fig:HART_UK_COMPAR_BOX}%
 		\end{subfigure}\hfill%
\begin{subfigure}{.52\columnwidth}
 			\includegraphics[width=1\columnwidth]{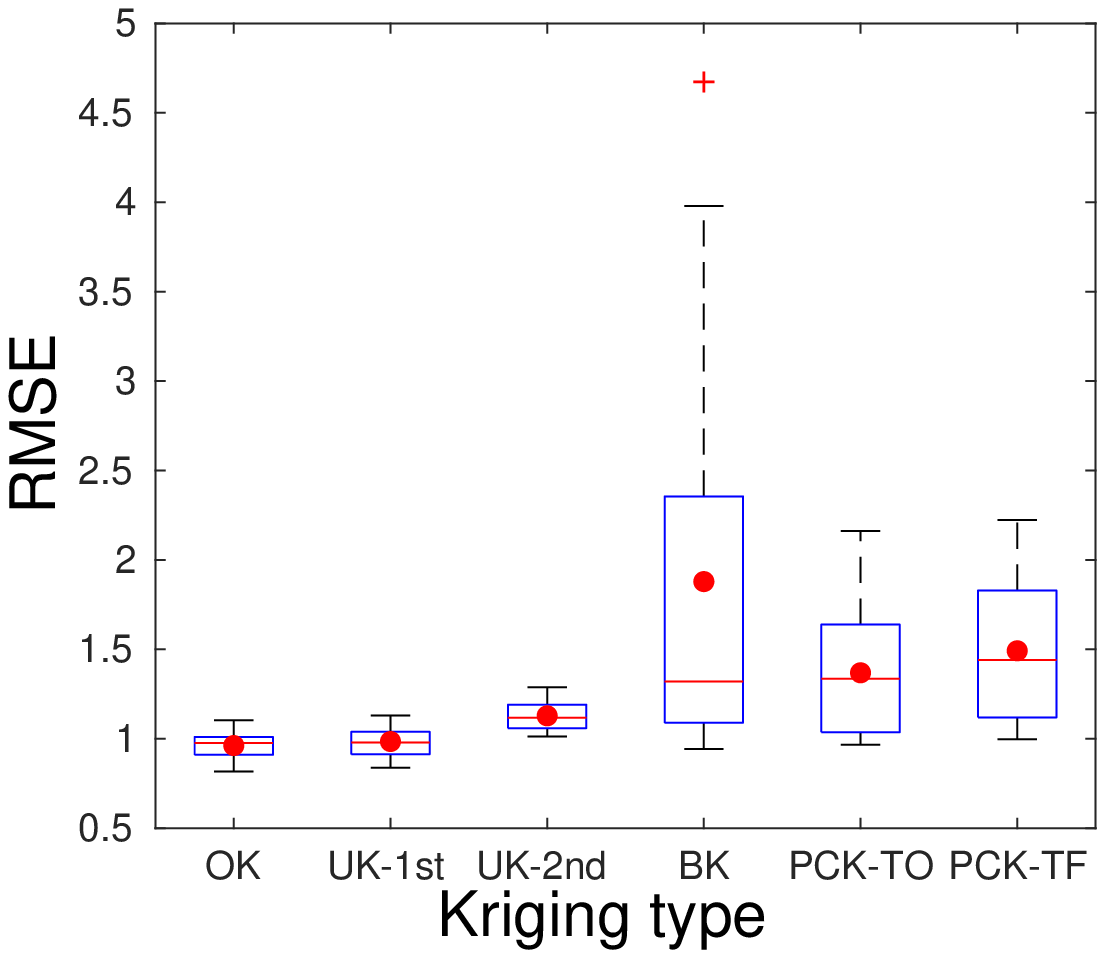}%
 			\caption{}%
 			\label{fig:HART_ERROR_PLOT_INT}%
 		\end{subfigure}\hfill%
 		\caption{Results obtained from various EGO schemes for the Hartman-6 function: \textbf{a} Convergence of the median, \textbf{b} boxplot, and \textbf{c} RMSE from the initial surrogate models.}
 		\label{fig:HARTplot}
 	\end{figure}

\begin{figure}
 		\centering
 		\begin{subfigure}{.5\columnwidth}
 			\includegraphics[width=1\columnwidth]{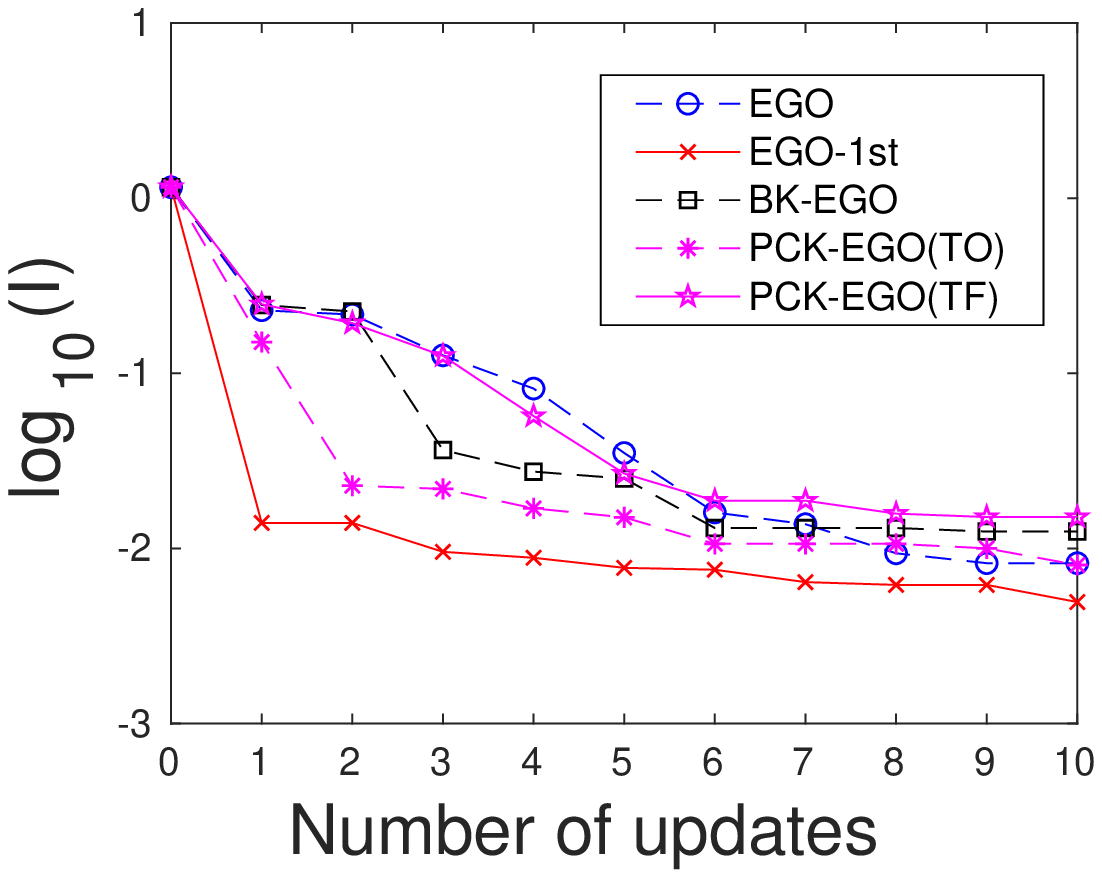}%
 			\caption{}%
 			\label{fig:BOREHOLE_UK_COMPAR}%
 		\end{subfigure}\hfill%
 		\begin{subfigure}{.5\columnwidth}
 			\includegraphics[width=1\columnwidth]{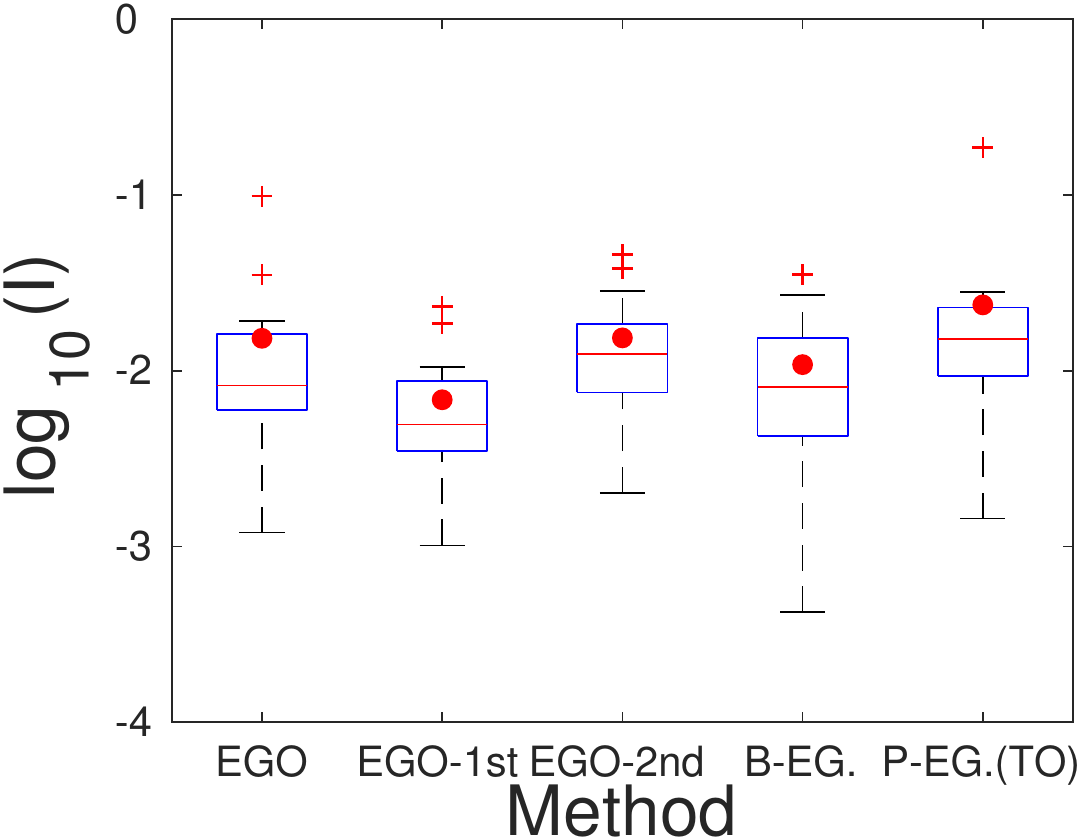}%
 			\caption{}%
 			\label{fig:BOREHOLE_UK_COMPAR_BOX}%
 		\end{subfigure}\hfill%
\begin{subfigure}{.5\columnwidth}
 			\includegraphics[width=1\columnwidth]{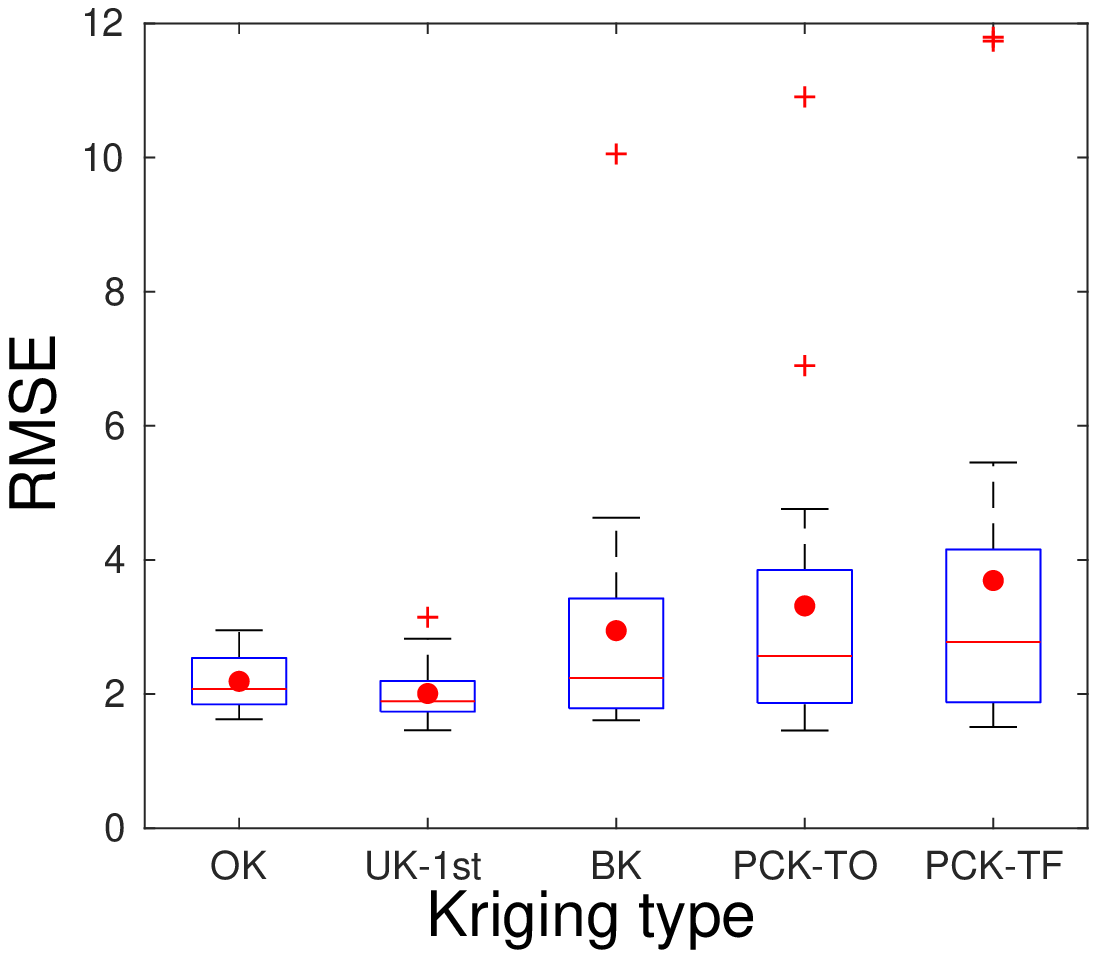}%
 			\caption{}%
 			\label{fig:BOREHOLE_ERROR_PLOT_INT}%
 		\end{subfigure}\hfill%
 		\caption{Results obtained from various EGO schemes for the borehole problem: \textbf{a} Convergence of the median, \textbf{b} boxplot, and \textbf{c} RMSE from the initial surrogate models.}
 		\label{fig:BOREHOLEplot}
 	\end{figure}

	We present our optimization results  in Figures~\ref{fig:braninplot}-\ref{fig:BOREHOLEplot}. For all test functions, we depicted median $I$ at each update of the optimization process and the boxplot of $I$ at the end of each search update. The plot of median convergence essentially depicts the capability of the optimizer in approaching the true optimum value of the function. Here, the faster the decay of $I$, the faster the optimizer locates the true optimum. On the other hand, the boxplots show the quality of solutions at the final iteration. As such, there is the possibility that an optimization algorithm scheme has a slow convergence of $I$, but is able to identify a satisfactory solution at the end of the search, or vice versa. Therefore, it is important to analyze these two aspects to obtain a complete understanding of the performance of the various optimization methods that we considered in this paper. We also depict the RMSE of the initial surrogate models obtained from OK and various UK schemes for all functions.
	
	For convenience, we denote EGO with OK, UK with first-order polynomial (i.e., UK-1st), UK with second-order polynomials (i.e., UK-2nd), BK, and PCK as EGO, EGO-1st, EGO-2nd, BK-EGO, and PCK-EGO, respectively. For the higher-dimensional problem, we also compared PCK that uses total-order expansion (i.e., PCK-TO) and maximum two-factor interactions (i.e., PCK-TF). We denote the PCK-EGO with total-order expansion and tensor product with maximum two-factor interactions as  PCK-EGO(TO) and PCK-EGO(TF), respectively. Note that, for the boxplot results of the Hartman-6 and borehole problems, to further improve the readability, we abbreviated BK-EGO, PCK-EGO(TO), and PCK-EGO(TF) to B.EG., P.EG(TO), and P.EG(TF), respectively. For the borehole problem we did not include the UK with second order polynomial since the polynomial size exceeded the sample size for this problem.
 		
 	For the two-dimensional functions, in general, both BK-EGO and PCK-EGO exhibited fast convergence of median $I$ relative to standard EGO. This is evident for the Branin and Hosaki functions in which both variants of UK-EGO with automatic trend function selection outperformed all other schemes. The use of first-order polynomials as trend functions was not sufficient to assist EGO for the Branin and Hosaki function; for the Hosaki function this worsened the optimization process. EGO-2nd was particularly useful for the Branin function but not for the Hosaki function, although its performance was still lower than both BK-EGO and PCK-EGO. The observed convergence behavior was more complex for the Sasena function; analysis of boxplots reveals that there was no significant difference between the qualities of final solutions for EGO, EGO-1st, BK-EGO, and PCK-EGO, with EGO-2nd emerging as the clear winner for the Sasena function. This result is in stark contrast with results obtained for the Branin and Hosaki functions in which BK-EGO and PCK-EGO surpassed the other methods. Even so, the fact that BK-EGO and PCK-EGO exhibited faster convergence versus that of EGO on early updates indicates that there was still a beneficial effect observed when UK with automatic trend function selection was employed for the Sasena function.
 	
 	For the Hartman-6 function, it was very difficult to achieve a very low $I$ value due to the highly nonlinear and complex nature of the response surface. We observe that all UK-EGO schemes were unable to perform better than standard EGO for the Hartman-6 function. Results for the Hartman-6 function indicate that the addition of a trend to model the surface of a function with no polynomial-like trend produced a poorer optimized solution compared to standard OK. For the borehole problem, even though the statistics for the final solutions show that there was no significant difference between EGO and PCK-EGO(TO), we observe that BK-EGO and PCK-EGO(TO) were faster than EGO in discovering locations near the optimum point. The fact that BK-EGO became quite stagnant near the end of the search indicates that it was unable to further exploit the global optimum of the borehole function. One fact worth noting is that EGO with a first-order polynomial outperformed the other methods starting from the very first iteration until the end of the search for the borehole problem. Therefore, using a first-order polynomial proves to be more than sufficient to improve the performance of UK-EGO relative to standard EGO for the borehole problem. This suggests that using automatic trend function selection does not always automatically lead to the best optimization performance. 
 	
 	On the Hartman-6 and borehole problem, the poor performance of PCK-EGO(TF) indicates that the initial candidate trend function also had a profound effect on the performance of PCK-EGO. Since BK-EGO and PCK-EGO(TF) use the same index of the candidate set, performance differences between the two might be attributed to the different types of polynomials and the trend function selection method.	
 	
	The RMSE results from the initial surrogate models show that the use of UK with automatic trend function selection (i.e., BK and PCK) does not automatically lead to improvements in approximation quality. For the Branin function in which the problem's surface can be well approximated with a polynomial, PCK demonstrated significant improvements in terms of accuracy; however, other functions featured a non-polynomial-like surface that posed a challenge for a polynomial-based response surface as that of the UK. We observe that sometimes using UK with a predefined trend function (i.e., first- or second-order polynomial) can produce a better approximation than that produced by BK and PCK, as in the case of Sasena, Hartman-6, and borehole problem. In high-dimensional problems, we also observe that the PCK-TO performed
	better than that of the PCK-TF.
	  	
 	Analyzing the relationship between the RMSE and the optimization performance, we observe that, in general, achieving a lower RMSE is closely correlated to better optimization performance. This is clear on such problems as Branin, Sasena, and borehole function in which PCK-EGO, EGO-2nd, and EGO-1st exhibited the best optimization performance, respectively. Especially in the Branin function, the relationship between the RMSE and the optimization performance is very clear. The RMSE is also useful for excluding poor performed surrogate models in terms of optimization, as it can be observed in the Hartman-6 function. However, as demonstrated in the Hosaki function, this was not always the case; the association between RMSE for the initial samples and optimization performance is not as clear for the Hosaki function. The fact that both BK-EGO and PCK-EGO performed better than EGO on the Hosaki problem in spite of a nonstatistically significant difference of RMSE results indicates that BK-EGO and PCK-EGO can better predict the global optimum location of the Hosaki function as compared to OK. Therefore, surrogate models that are less accurate in the entire design space can produce better-optimized solutions than OK, thus indicating that they can better predict where the optimum lies.

 	\subsection{Studies in aerodynamic design optimization}	
 	Having finished our study on synthetic problems, we next focus our analysis on the capabilities of UK-EGO in real-world optimization problems. The problem we consider in this paper is transonic airfoil optimization in inviscid flow using class shape transformation (CST) airfoil parameterization~\citep{kulfan2008universal} and an Euler CFD solver. This problem was previously studied by~\cite{ray2004swarm} in the context of multiobjective optimization and by~\cite{palar2016comparative} in the context of a real-world application of multiobjective surrogate-based memetic algorithms. 
 	
\begin{figure}
 		\centering
 		\begin{subfigure}{.49\columnwidth}
 			\includegraphics[width=1\columnwidth]{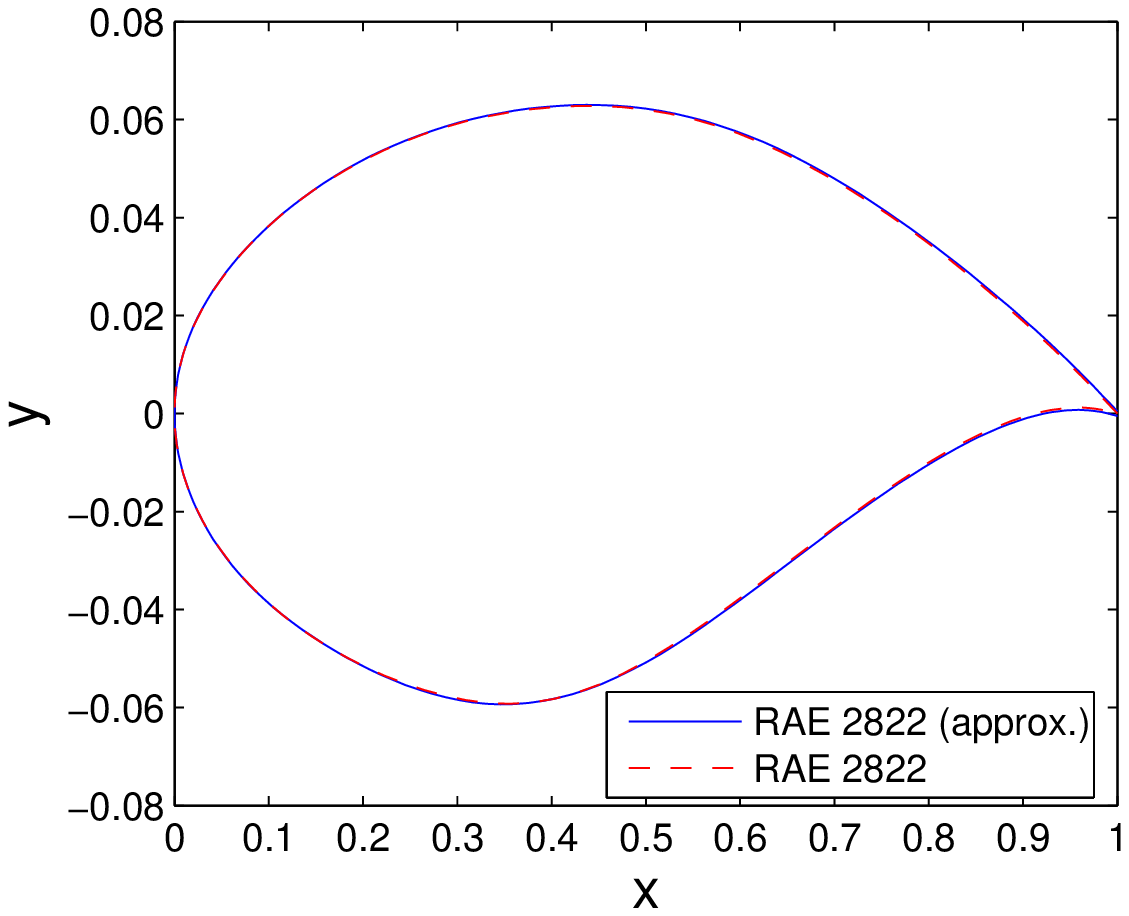}%
 			\caption{}%
 			\label{fig:RAE2822_CST}%
 		\end{subfigure}\hfill%
 		\begin{subfigure}{.49\columnwidth}
 			\includegraphics[width=1\columnwidth]{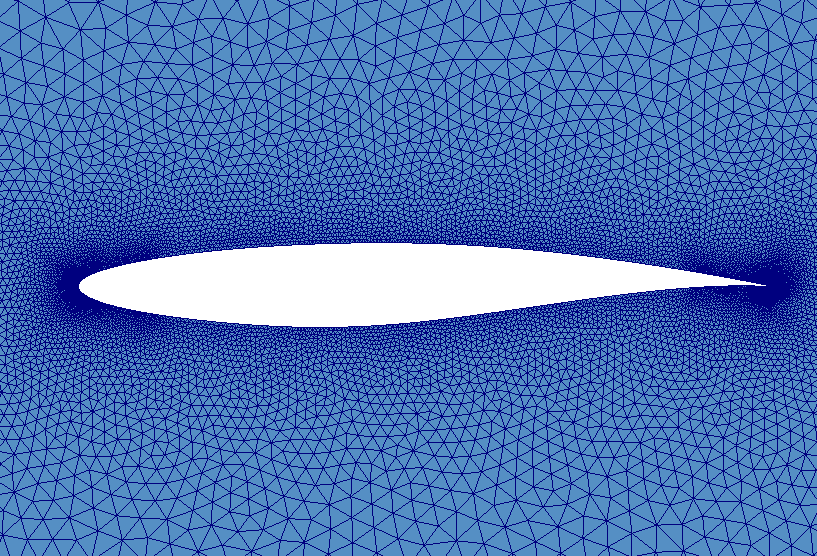}%
 			\caption{}%
 			\label{fig:RAE2822_MESH}%
 		\end{subfigure}\hfill%
 		\caption{Baseline airfoil used in this paper: \textbf{a} original RAE 2822 and CST approximation and \textbf{b} inviscid mesh for the CFD.}
 		\label{fig:RAE2822}
 	\end{figure}

 	For this specific real-world problem, we used low-fidelity inviscid Euler code to obtain the aerodynamic coefficients. Although this inviscid solver is rarely used in real-world airfoil optimization problems, it is representative of this real-world aerodynamic optimization problem in terms of the response surface's complexity. Moreover, the use of a low-fidelity solver allows us to collect results from several optimization runs to then perform a statistical analysis of our results. The objective here is to minimize the ratio of the drag-to-lift coefficient ($C_{d}/C_{l}$) as a function of CST shape parameterization. We used a 16 variable CST to represent the airfoil shape and therefore act as decision variables. To set the upper and lower bounds for optimization, we first fit the RAE 2822 airfoil geometry with CST parameterization to identify the initial CST parameters. These initial parameters were then varied within $\pm20\%$ of their initial values to serve as decision variables. Here, the design conditions for optimization are $M=0.8$ and $AoA=2^{\degree}$, where $M$ and $AoA$ are the Mach number and angle of attack, respectively. The number of initial samples was 40 with 15 additional samples, where LHS was used to generate 10 different sets of initial samples. We intentionally set the initial sample size to a low value (i.e., $2.5m$=40) to simulate a high-dimensional optimization problem with sparse initial sampling points, with an additional 15 enriched samples to seek the optimum solution. The enlarged geometry and CFD mesh of the RAE 2822 airfoil used in our simulation are shown in Fig~\ref{fig:RAE2822}. Here, the value of $C_{d}/C_{l}$ obtained from CFD simulation for the datum airfoil was 0.0732. 
 	
 	Next, we set two values of $p_{max}$ to construct BK and PCK for this problem, i.e., $p_{max}=1$ and $p_{max}=2$. We also compared our results with a PCK-EGO that uses maximum two-factor interactions for trend function generation. Moreover, we also compared our results with the UK that directly uses the coefficient magnitude to build an optimum UK surrogate model (i.e., the frequentist viewpoint)~\citep{couckuyt2012blind}, which we denoted as UK-1st(F). In this paper, the frequentist viewpoint ranks the polynomial terms according to their coefficients which are estimated via GLS. Following this step, multiple Kriging models are then constructed according to this ranking; the Kriging model of lowest LOOCV error is then selected. Clearly, this can only be done if the number of polynomial term is lower than the number of samples available. 
 	
\begin{figure}
 		\centering
 		\includegraphics[width=0.7\columnwidth]{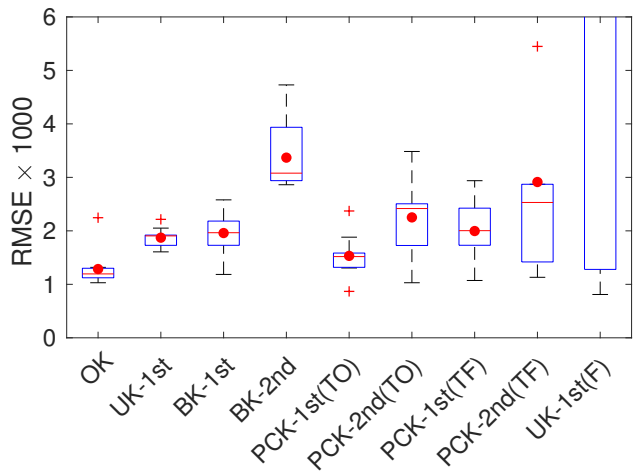}%
 		\caption{RMSE results from OK and various UK schemes on the airfoil problem.}%
 		\label{fig:AIRFOIL_UK_ERROR_2}%
 		
 	\end{figure}

	Firstly, using a different set of LHS generated initial samples and 200 validation samples, we analyzed the approximation quality of the various UK schemes in the given aerodynamic function. Our results, which we present in Figure~\ref{fig:AIRFOIL_UK_ERROR_2}, show that all choices of nonconstant trend function produced surrogate models with inferior quality as compared to the standard OK. Among UK variants, PCK with total-order expansion of $p_{max}=1$ produced the best approximation. PCK was also generally better than BK in approximating the aerodynamic function. Further, PCK surrogate models with total-order expansion were also better than PCK with maximum two-factor interaction in terms of approximation error. The fact that the approximation worsened again for all types of BK and PCK when $p_{max}$ was set to two indicates that higher values of $p_{max}$ does not ensure a better approximation quality. This essentially means that the LOOCV scheme failed to detect the proper trend function for the UK approximation; thus, it is probably better to limit $p_{max}$ to avoid poor approximation. Nonetheless, all UK variants with automatic trend function selection produced-higher quality surrogate models versus that of UK-1st(F). These results show that using the automatic trend function selection procedure to build a UK is more beneficial than directly using the order from the coefficients magnitude to approximate this problem's landscape. By considering the fact that we have a similar trend function set for UK-1st, PCK-1st(TO), and UK-1st(F), the low RMSE value of PCK-1st(TO) can be attributed to the success of the LARS algorithm itself, though we note that the RMSE was still higher than that of OK.
 	
\begin{figure}
 		\centering
 		\includegraphics[width=0.75\columnwidth]{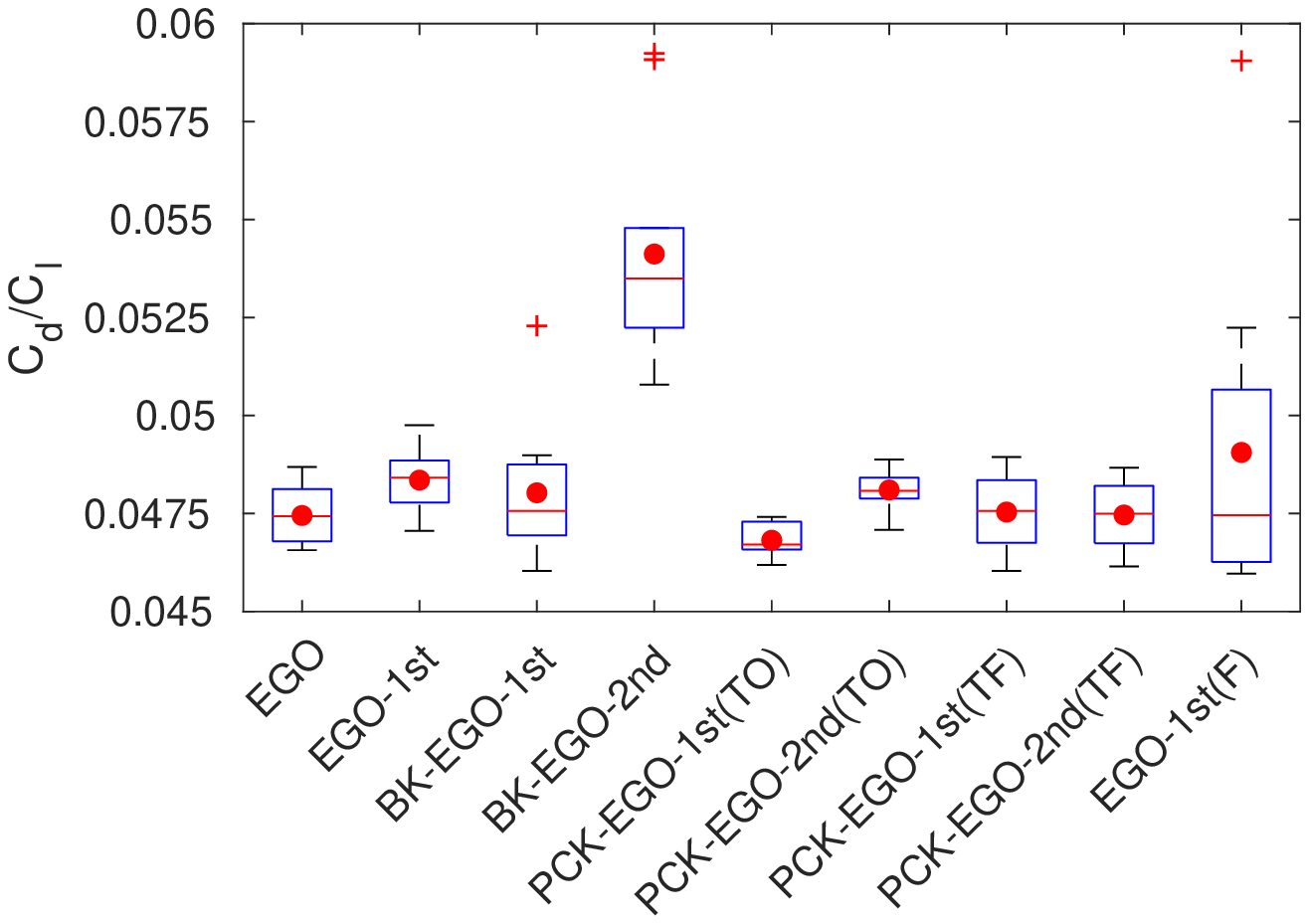}%
 		\caption{Boxplot of the optimum solutions at the final iteration on the airfoil problem.}%
 		\label{fig:AIRFOIL_UK_BOX}%
 	\end{figure}

 	In further explaining our results, we separate the convergence plots into two figures to avoid a too convoluted view. We also separate this plot such that it is easier for us to investigate the effects of different UK methods (i.e., BK and PCK) and the benefit of automatic trend function selection as compared to UK with fixed polynomial trend functions. Note that Figures~\ref{fig:AIRFOIL_UK_COMPAR_MEAN_BARU} and~\ref{fig:AIRFOIL_UK_COMPAR_MEAN_BARU_2} show the convergence plot of the mean value and not the median; we use this approach here because we only performed our experiments 10 times due to the costly function evaluation; hence plotting the mean makes more sense. The boxplot of best solutions at the end of the search are depicted in Fig.~\ref{fig:AIRFOIL_UK_BOX}.
 	
 	\subsubsection{Comparing blind Kriging with polynomial-chaos-Kriging and the effect of $p_{max}$ }
\begin{figure}
 		\centering
 		\includegraphics[width=0.65\columnwidth]{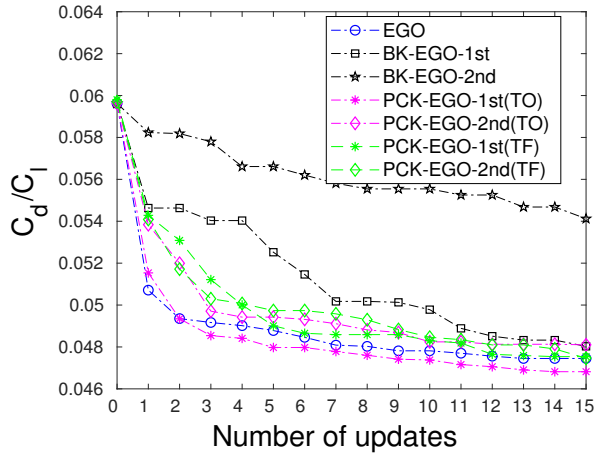}%
 		\caption{Mean convergence of the optimum solutions for the airfoil problem with comparisons to BK and PCK.}%
 		\label{fig:AIRFOIL_UK_COMPAR_MEAN_BARU}%
 	\end{figure}

 	We first analyze and compare the performance of BK-EGO and PCK-EGO relative to standard EGO. We also analyze the effect that $p_{max}$ had on the performance of PCK-EGO and BK-EGO, as well as the effect that candidate trend function set had on PCK-EGO. Our results are shown in Figure~\ref{fig:AIRFOIL_UK_COMPAR_MEAN_BARU}; we observe that a proper choice of trend function for UK can improve the performance of our EGO-based optimization. Here, PCK-EGO(TO) with $p_{max}=1$ was the best performer for the airfoil problem surpassing EGO with OK. The success of PCK-EGO(TO) with $p_{max}=1$ here indicates the linearity or almost linear behavior of relationship between the objective function and decision variables. More specifically, PCK-EGO-1st(TO) was the only UK variant that outperformed standard EGO with OK for this problem. 
 	
 	We also observe here a significant difference between performances of BK-EGO and PCK-EGO, where all variants of the latter outperformed the former. We also observe a notable effect when total-order expansion was used to build the candidate trend function for PCK-EGO. This trend of better performance for PCK-EGO with total-order expansion is similar to our results for the borehole and Hartman-6 problems. We note here that PCK-EGO achieved its optimum performance on a high-dimensional problem if the candidate trend function set was constructed through total-order expansion, that allowed interactions of higher than two factors to be included in the scheme. The poor performance of BK-EGO here can be attributed to the polynomial trend function type, trend function selection scheme, and most likely the choice of the candidate trend function set. 
 	
 	Finally, the effect of $p_{max}$ was profound in this example, with the performance of BK-EGO and PCK-EGO deteriorating when the value of $p_{max}$ was set to two. This trend indicates that applying a candidate polynomial set with a higher order does not automatically lead to a better optimization process, even with an automatic trend function selection procedure. This is also true when we want to find a better-optimized solution versus that of the one identified by standard EGO with OK. Here, the success of PCK-EGO-1st(TO) can be attributed to at least two factors, that is, the LARS algorithm and the correct choice of the candidate trend function set. To investigate whether the LARS algorithm yielded better performance for PCK-EGO-1st(TO) as compared to EGO with OK, we further compared the performance of PCK-EGO-1st(TO) with EGO with all first order polynomial functions and the frequentist viewpoint.

 	\subsubsection{Comparing polynomial-chaos-Kriging with Universal Kriging based on a frequentist viewpoint and fixed trend}
\begin{figure}
 		\centering
 		\includegraphics[width=0.65\columnwidth]{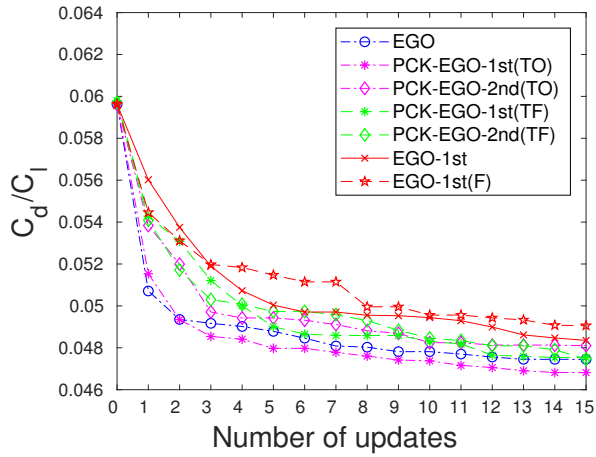}%
 		\caption{Mean convergence of the optimum solutions for the airfoil problem with comparisons to UK-EGO with a fixed trend function and frequentist ordering.}%
 		\label{fig:AIRFOIL_UK_COMPAR_MEAN_BARU_2}%
 	\end{figure}

 	Our results are shown in Figure~\ref{fig:AIRFOIL_UK_COMPAR_MEAN_BARU_2}. For clarity, we again show the performance of all PCK-EGO results with total-order expansion and maximum two-factor interactions. Here, we observe that PCK-EGO with LARS had better performance as compared to the PCK-EGO with all first-order polynomial functions or with frequentist trend function ordering. When building the intermediate UK surrogate models, the LARS algorithm considers the correlation between the residuals from the kernels. This is the primary difference between PCK with LARS and the one with a simple frequentist scheme, that does not consider this residual when building the intermediate surrogate models. In this sense, we can see that, for the aerodynamic problem, the LARS algorithm successfully selected the important first-order polynomial that was then able to guide the optimization process to discover the optimum solution even when the cardinality of the candidate set was lower than the sample size.
 	
 	If we again observe the RMSE plot in Figure~\ref{fig:AIRFOIL_UK_ERROR_2}, there is a clear correlation between the RMSE and the optimization performance for the airfoil problem. However, there are some exceptions, the extremely poor optimization performance of BK-EGO-2nd was not observed in the RMSE plot, since the mean and median RMSE values of BK-2nd were still lower than those of UK-1st(F) in which the EGO-1st(F) exhibited better performance than that of BK-EGO-2nd. Moreover, PCK-EGO-1st(TO), which was the best performer for this problem, had even higher mean and median RMSE values as compared to standard OK. 
 	
 	\subsection{Remarks from the test problems}
 	Based on the results of the tests on the synthetic and real-world problems, we can make several remarks and recommendations regarding the application of UK for EGO-based optimization:
 	\begin{enumerate}

 	\item \textbf{The PCK-EGO was the best variant among all UK-based EGO methods.} Furthermore, our investigation showed that EGO with UK had the highest performance when the landscape of the problem exhibited a trend that could be effectively captured by a polynomial. The UK variants with automatic trend function selection (i.e., BK and PCK) were able to outperform both OK and UK with a fixed trend function on problems exhibiting polynomial or slightly nonlinear trends, such as the Branin and Hosaki functions. Meanwhile, all UK-EGO variants failed to outperform standard EGO for problems with highly nonlinear trends and problems in which no clear polynomial trend exists (e.g., the Hartman-6 function). 
 	\item \textbf{For high-dimensional problems, using total-order expansion is the best choice when one wishes to apply PCK-EGO.} Our results signify that there is a potential for improvement when UK is used for EGO-based optimization, albeit with the warning that UK with automatic trend function selection does not automatically lead to a better optimization process. 
 	\item \textbf{UK surrogate models with evidently lower and higher approximation error versus that of OK indicates a good and poor surrogate for EGO, respectively}. However, UK surrogate model with similar or slightly lower quality versus that of OK can also perform better or at least similar to the OK when applied within EGO algorithm. This is mainly due to the addition of trend function that aids the process of locating the optimum point by providing a good direction for optimization. 
 	\item \textbf{Although a proper choice of trend function for UK can result in better-optimized solutions, care should be taken when employing UK in solving real-world problems}. It is obvious that an improper selection of trend function can result in an inferior optimized solution as compared to standard OK. Note that this is also true when UK with automatic trend function selection is employed. We observe that when addressing real-world optimization, PCK-EGO with $p_{max}=1$ and total-order expansion was the best variant of all, whereas increasing $p_{max}$ only worsened performance as compared to that of OK.
	\end{enumerate}
	
 	Based on points 1 and 2, we therefore recommend the use of PCK-EGO with total-order expansion-generated polynomial trend for solving expensive single-objective optimization problems. We have observed so far that this method ensures a robust performance of EGO-based search when seeking the optimum solution as compared to other UK-EGO algorithms. Regarding points 3 and 4, we recommend carefully performing a preliminary analysis of the problem's complexity before solving the problem. If possible, one should analyze the degree of nonlinearity and interaction between variables in the problem being tackled. Note that UK with automatic basis selection (i.e. BK and PCK) provides a tool for such interpretation from the given trend function. Nonetheless, our personal recommendation is to use a variance-based sensitivity analysis tool (Sobol method) since it can measure the effect of interactions besides the main effect~\citep{sobol1993sensitivity}. As suggested by~\cite{couckuyt2012blind}, using an independent test set is also helpful. For example, one can construct a UK model with this independent test set to compare the error performance of various UK and OK implementations, for additional information besides the LOOCV error. We believe that these are important steps for successful implementation of UK-EGO algorithms. 
 	
    
 	
 	\hfill%
 	
 	\section{Conclusions}
 	\label{sec:5}
 	In this paper, we investigated the capabilities of UK in assisting optimizations using the EGO framework. Our primary objective was to analyze the potential benefits and pitfalls of UK when used to solve expensive simulation-based optimization problems; secondarily, we investigated the computational aspects of UK with automatic trend function selection. We first explained the UK surrogate model and EGO, stressing the automatic trend function selection methods for UK, which are BK and PCK.
 		
 	The capabilities of UK coupled with EGO were then investigated via five test problems and one real-world engineering (i.e., aerodynamic) problem. More specifically, we studied four variants of UK, each of which was compared with OK: these were (1) UK with a first-order polynomial, (2) UK with a second-order polynomial, (3) BK, and (4) PCK. EGO with PCK was the best variant tested, as it was able to perform better than or at least similarly to (or slightly worse than) EGO with OK. PCK-EGO was not necessarily the best for all test functions, but it was more robust than the other UK schemes. In general, PCK-EGO was better and more robust than BK in several instances of test problems; however, it proved to be much better to equip PCK-EGO with total-order expansion-generated candidate polynomial sets to ensure optimum performance for high-dimensional problems. For the real-world problem, determining the candidate polynomial set for automatic  trend function selection also needed careful consideration, since PCK-EGO was able to perform better than EGO with OK only with a proper choice of its candidate polynomial set. Based on our investigations, a UK surrogate model that is more accurate than OK in its initial iteration was able to produce an optimized solution with better quality; however, we also found that a surrogate model that is less accurate in modeling the entire design space was able to produce a better-optimized solution versus that of OK. This phenomenon occurred because in spite of its global accuracy, 
 	this approach was able to point out the locations of the global optimum better than standard OK.
 	
 	Although in this paper we only considered unconstrained functions, extending our work to constrained problems is relatively straightforward; here, the constraint surrogate model would also be constructed using the UK surrogate model. As for our future work, we plan to develop other criteria aside from just LOOCV error identification when choosing a suitable UK surrogate model for EGO. We include this as future work because LOOCV error favored surrogate models that were globally accurate, whereas for optimization, a surrogate model that is locally accurate near the optimum region is more desirable. We believe that more real-world studies are needed to further investigate the capability of UK-EGOs in solving various real-world problems. Finally, there is potential for parallelization since both BK and PCK construct multiple Kriging models that can be exploited for such task.

 	 \section*{Appendix A: Trend function selection and hyperparameter optimization strategy for PCK}
 	  
 	  We propose a sequential hyperparameter optimization strategy based on BFGS at each iteration of the UK construction process that utilizes the optimum solution from the previous iteration. This strategy can be applied to both BK and PCK since both methods work by scanning from the provided polynomial set. Regardless, our strategy uses the optimum solution obtained in the previous iteration as the initial solution for the BFGS search in the current iteration. Our primary motivation for applying this strategy is that the likelihood function for one iteration might change only slightly relative to the previous iteration, indicating the proximity of the global optimum hyperparameter locations. Further, in our strategy, a GA is only used in the first iteration to find the optimum hyperparameters of the OK before adding more trend functions. Here, we utilize a GA with a population size of 100 and a maximum of 200 generations followed by BFGS search. This exhaustive search is used only in the first iteration since the accuracy of the hyperparameters' optimization procedures that follow relies on the accuracy of the OK hyperparameters. After the final trend function is identified, our GA+BFGS approach is then applied again with this final trend function to search for possible higher values of the likelihood. We call our strategy here the simplified GA+BFGS strategy as opposed to the exhaustive GA+BFGS strategy. 
 	 	 	
 	 To verify the performance of our simplified GA+BFGS strategy, we compared its performance with the exhaustive GA+BFGS strategy using five test functions mentioned in Appendix B.  In this study, we set sample size to 20, 60, and 40 for the two-dimensional problems, Hartman-6, and borehole problem, respectively. We generally used $N_{s}=10\times m$, where $N_{s}$ is the sample size, to generate the sample set, with the only exception being the borehole problem in which we set the sample size to 40 to make the optimization problem more difficult.  We also compared our simplified GA+BFGS strategy with the simple BFGS strategy that employs a one-shot strategy with a random initial solution at each UK iteration.  More specifically here, we compared the lowest LOOCV errors resulting from these three strategies. 
 	 
 	  	 	 	\begin{figure}
 	  	 	 		\centering
 	  	 	 		\begin{subfigure}{.49\columnwidth}
 	  	 	 			\includegraphics[width=1\columnwidth]{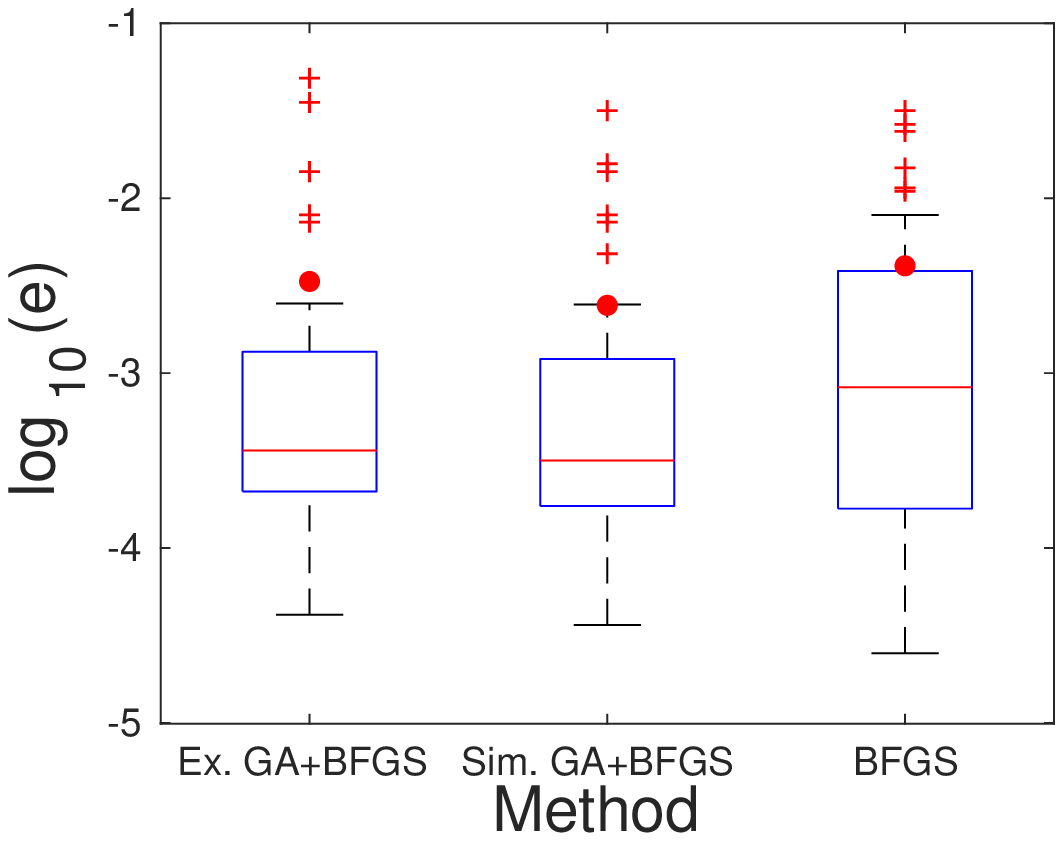}%
 	  	 	 			\caption{}%
 	  	 	 			\label{fig:LOOBOX_BRA}%
 	  	 	 		\end{subfigure}\hfill%
 	  	 	 		\begin{subfigure}{.49\columnwidth}
 	  	 	 			\includegraphics[width=1\columnwidth]{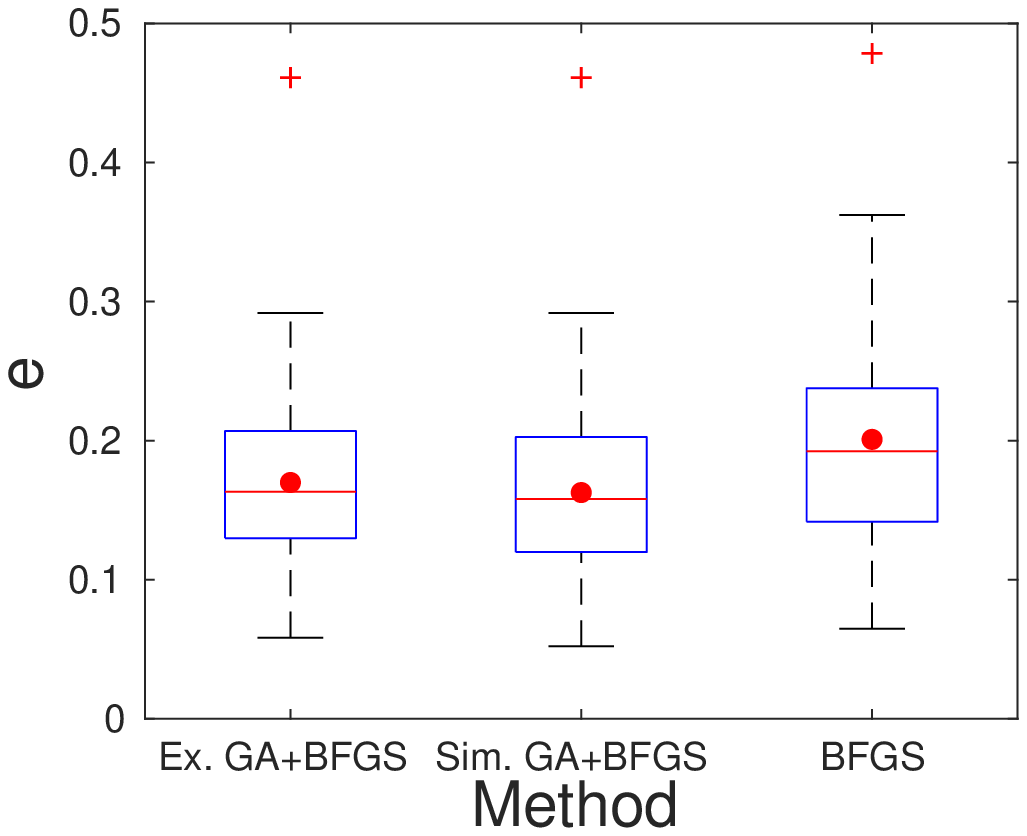}%
 	  	 	 			\caption{}%
 	  	 	 			\label{fig:LOOBOX_SAS}%
 	  	 	 		\end{subfigure}\hfill%
 	  	 	 		\begin{subfigure}{.49\columnwidth}
 	  	 	 			\includegraphics[width=1\columnwidth]{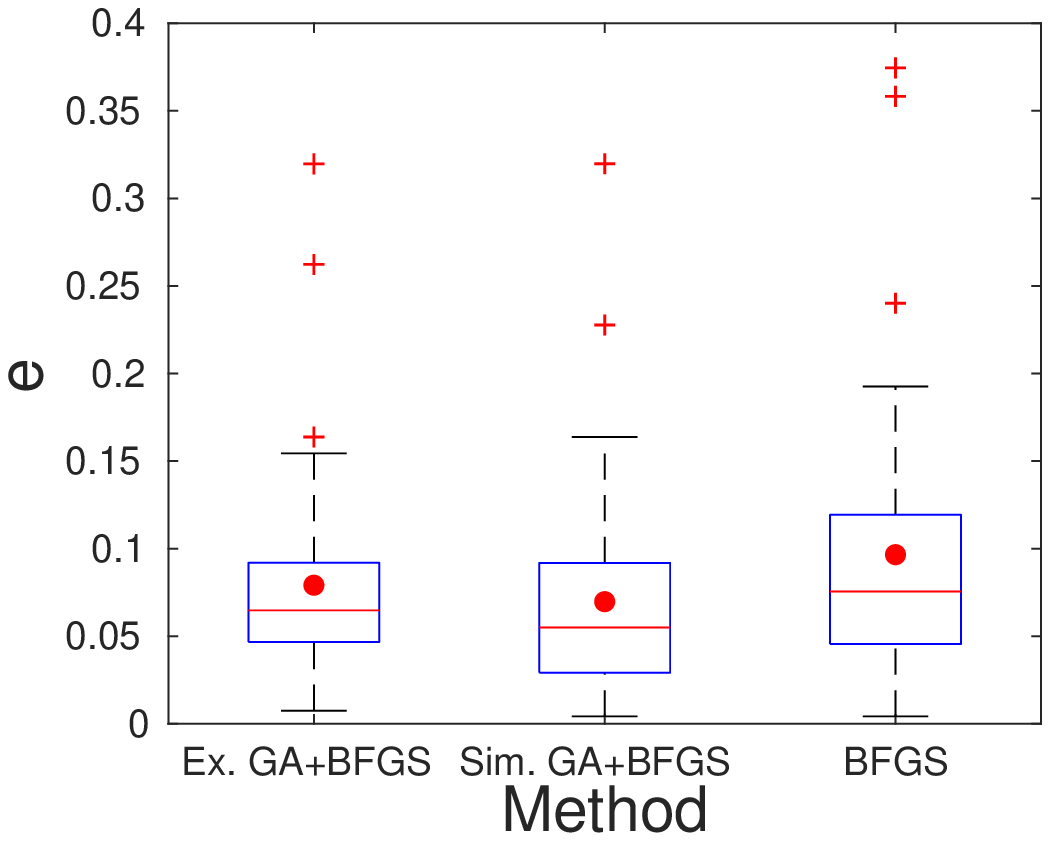}%
 	  	 	 			\caption{}%
 	  	 	 			\label{fig:LOOBOX_HOS}%
 	  	 	 		\end{subfigure}\hfill%
 	  	 	 		\begin{subfigure}{.49\columnwidth}
 	  	 	 			\includegraphics[width=1\columnwidth]{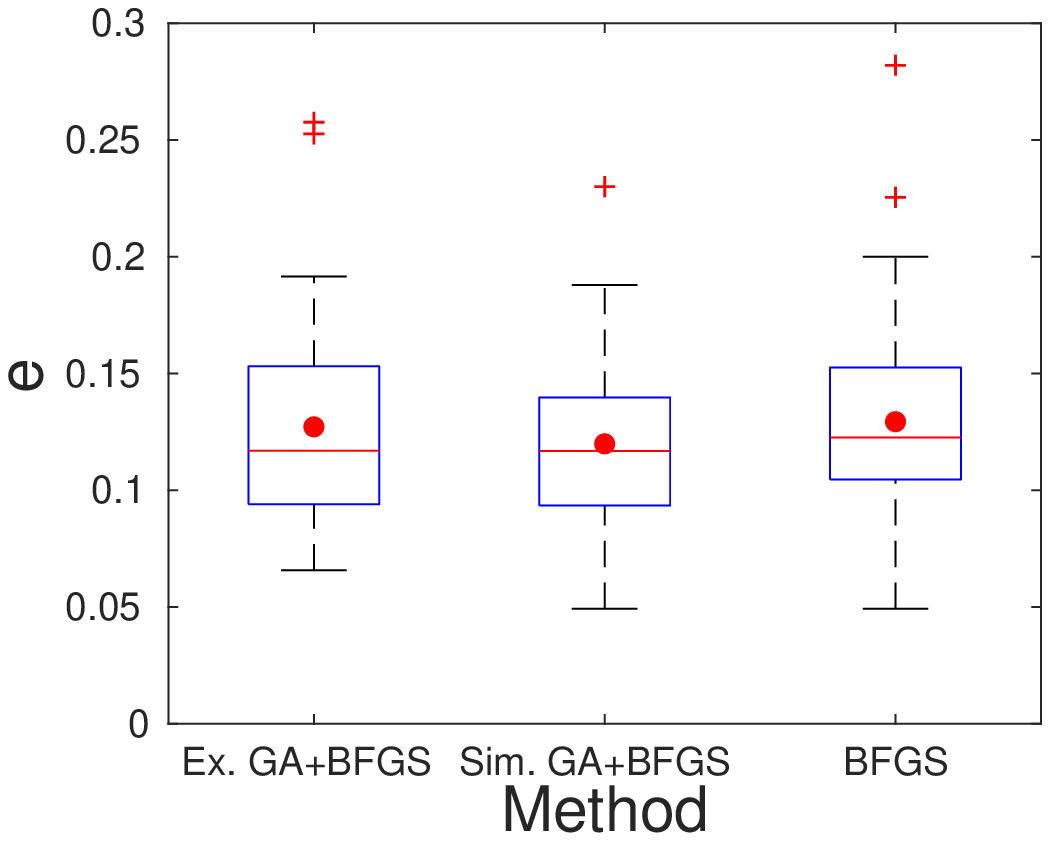}%
 	  	 	 			\caption{}%
 	  	 	 			\label{fig:LOOBOX_HART}%
 	  	 	 		\end{subfigure}\hfill%
 	  	 	 		\begin{subfigure}{.49\columnwidth}
 	  	 	 			\includegraphics[width=1\columnwidth]{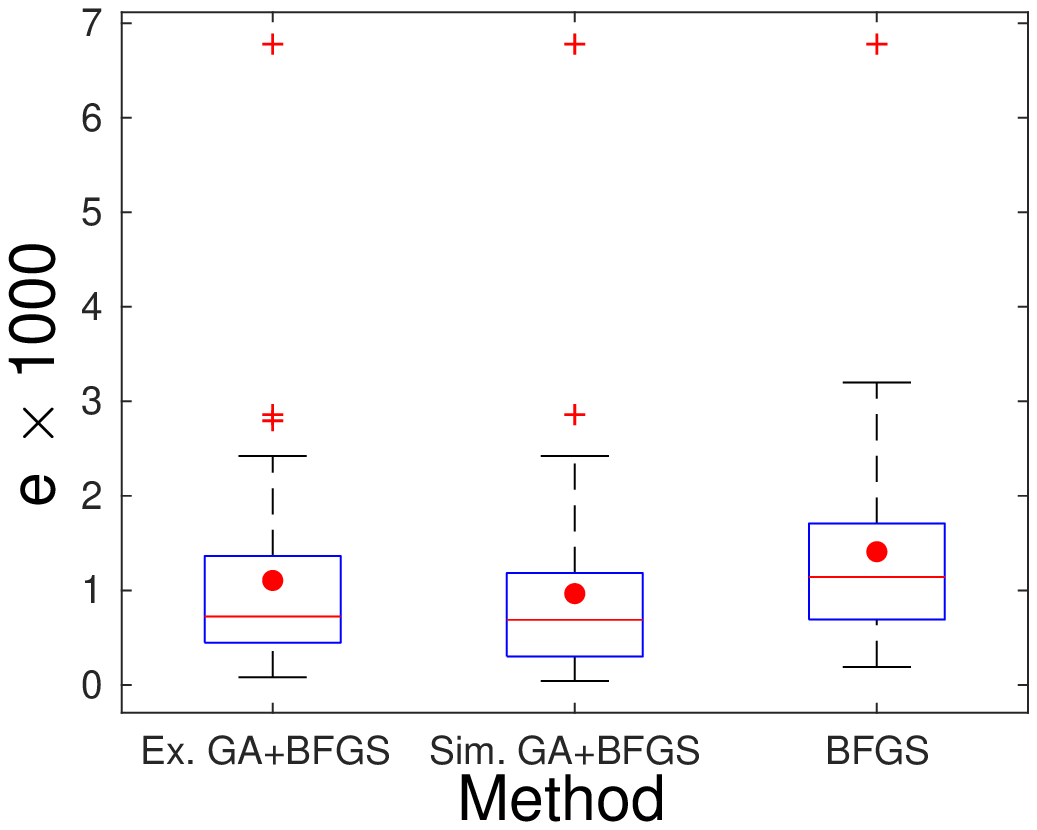}%
 	  	 	 			\caption{}%
 	  	 	 			\label{fig:LOOBOX_BOR}%
 	  	 	 		\end{subfigure}\hfill%
 	  	 	 		
 	  	 	 		\caption{Comparing the LOOCV error for PCK with hyperparameters tuned using various strategies on the:\textbf{a.} Branin, \textbf{b.} Sasena, \textbf{c.} Hosaki, \textbf{d.} Hartman-6, and \textbf{e.} borehole problems.}
 	  	 	 		\label{fig:LOOBOX}
 	  	 	 	\end{figure}
 	 	 	 	
 	 For all five test functions, we observe from Figure~\ref{fig:LOOBOX} that the error performances of the UK for the exhaustive GA+BFGS and simplified GA+BFGS strategies are similar to one another. We observe that the performance of the simple BFGS strategy was not as good as that of the other two strategies. All strategies performed approximately the same for the Hartman-6 problem; this problem is a highly nonlinear and difficult problem in which the UK did not perform better than the standard OK in terms of approximation quality, thus explaining why UK hyperparameter tuning minimally affects LOOCV error. The lower performance of the simple BFGS strategy here signifies that the discovery of the optimum of a likelihood function for the UK is sensitive to the choice of the initial point. 
 	 	 	 	
 	 The time required to train the hyperparameters using these simplified and exhaustive strategies on a two-dimensional function with $p=4$ was approximately 3 and 40 seconds, respectively, on a computer with Intel$^{\textregistered}$ Xeon(R) E5-1630 v4 8 core CPU @ 3.70GHz equipped with MATLAB. This indicates that our simplified strategy can perform similarly to the exhaustive strategy in only 7.5\% of the time required by the exhaustive approach. 
 	 
 	\section*{Appendix B: Test functions}
 	\label{sec:appendixB}
 	\begin{enumerate}
 		\item Branin function (two variables).
 	\begin{equation}
 	f_{1}(\boldsymbol{x}) = \bigg(b_{2}-\frac{5.1}{4\pi^{2}}b_{1}^{2}+\frac{5}{\pi}b_{1}-6 \bigg)^{2} + 10 \bigg[\bigg(1-\frac{1}{8\pi} \bigg) \mbox{cos }(b_{1})+1\bigg], \\
 	\end{equation}
 			
 	where $b_{1}=15x_{1}-5$, $b_{2}=15x_{2}$, and $x_{1},x_{2}\in[0,1]^{2}$. 
 		
 		\item Sasena function (two variables). 
 		\begin{multline}
 		f(\boldsymbol{x}) = 2 + 0.01(x_{2}-x_{1}^{2})^{2}+(1-x_{1})^{2} + 2(2-x_{2})^{2} + 7\text{sin }(0.5x_{1}) \text{sin }(0.7x_{1}x_{2}).  \\  
 		x_{1}\in[0,5], x_{2}\in[0,5]. \\
 		\end{multline}
 		\item Hosaki function (two variables)
 		\begin{multline}
 		f(\boldsymbol{x}) = \big(1-8x_{1}+7x_{1}^{2} - (7/3)x_{1}^{3} +(1/4)x_{1}^{4} \big) x_{2}^{2}e^{-x_{1}}. \\  
 		x_{1}\in[0,5], x_{2}\in[0,5].\\
 		\end{multline}
 		
 		\item Hartman-6 function (six variables)
 		
 		\begin{multline}
 		f(\boldsymbol{x}) = -\sum_{i=1}^{4}c_{i}\text{exp } \bigg\{-\sum_{j=1}^{n}\textrm{A}_{ij}(x_{j}-\textrm{P}_{ij})^2\bigg\} , \\
 		\boldsymbol{x}=(x_{1},x_{2},\ldots,x_{n})^{T}, x_{i}\in[0,1] \\ 
 		\end{multline}
 		
 		where
 		\begin{equation}
 		\boldsymbol{c} = [1.0, 1.2, 3, 3.2]^{T}.
 		\end{equation}
 		
 		\begin{equation}
 		\textrm{\textbf{A}} = 
 		\begin{bmatrix}
 		10& 3& 17& 3.5& 1.7& 8\\
 		0.05& 10& 17& 0.1& 8& 14\\
 		3& 3.5& 1.7& 10& 17& 8\\
 		17& 8& 0.05& 10& 0.1& 14\\
 		\end{bmatrix}
 		\end{equation}
 		
 		\begin{equation}		
 		\textrm{\textbf{P}} = 10^{-4} 
 		\begin{bmatrix}
 		1312& 1696& 5569& 124& 8283& 5886\\
 		2329& 4135& 8307& 3736& 1004& 9991\\
 		2348& 1451& 3522& 2883& 3047& 6650\\
 		4047& 8828& 8732& 5743& 1091& 381
 		\end{bmatrix}		
 		\end{equation}
 		
 		\item Borehole function (eight variables)
 		\begin{equation}
 		f(\boldsymbol{x}) = \frac{2\pi T_{u}(H_{u}-H_{l})}{\text{ln}(r/r_{w})\left(1+\frac{2LT_{u}}{\text{ln}(r/r_{w})r_{w}^{2}K_{w}}+\frac{T_{u}}{T_{l}}\right)}
 		\end{equation}
 		
 		where the input variables are defined as shown in Table~\ref{tbl:borehole}.
 		
 		\begin{table}[h]
 			\centering
 			\begin{tabular}{cc} \hline
 				Random variable & Uncertainty range \\ \hline 
 				$r_{w}$   & [0.05; 0.15]   \\ 
 				$r$  &   [100; 50000]  \\ 
 				$T_{u}$  &  [63700; 115600]  \\ 
 				$H_{u}$  &  [990; 1100]  \\ 
 				$T_{l}$   & [63.1; 116]   \\ 
 				$H_{l}$   &  [700; 820]  \\ 
 				$L$ &   [1120; 1680]  \\ 
 				$K_{w}$  &[9855; 12045]   \\ \hline
 			\end{tabular}
 			\caption{The input variables and their input ranges for the borehole test function.}
 			\label{tbl:borehole}
 		\end{table}
 	\end{enumerate}

 	\section*{Appendix C: Boxplot}
 	For the boxplots, the bottom and top of each box represent the lower quartile Q1 (i.e., 25\%) and upper quartile Q3 (i.e., 75\%), respectively. The line between the top and bottom of the box represents the median (i.e., 50\%). Further, the whiskers below and above the box are drawn from $Q1-1.5$ IQR and $Q3+1.5$ IQR, where IQR represents the interquartile range (i.e., Q3-Q1). Observations that lie beyond the whisker length are identified as outliers. Finally, the circle denotes the mean of the observations.

 	
 	
 	\nocite{*}
 	\bibliography{Bibliography}
 	
 \end{document}